%% file: main.tex
\documentclass[10pt,twocolumn,letterpaper]{article}

\PassOptionsToPackage{dvipsnames,svgnames}{xcolor}
\usepackage{cvpr}              

\input{math_commands.tex}

\input{preamble}

\definecolor{cvprblue}{rgb}{0.21,0.49,0.74}
\usepackage[pagebackref,breaklinks,colorlinks,allcolors=cvprblue]{hyperref}

\def\paperID{*****} 
\def\confName{CVPR}
\def\confYear{2026}

\title{\texorpdfstring{\LiveScaleLogo}{LiveAct}: Towards Real-Time Multi-Agent World Models\\with Heterogeneous Action Interfaces}

\author{%
  Xinyu Wang\textsuperscript{1},
  Huafeng Shi\textsuperscript{2},
  Yue Ma\textsuperscript{3},
  Zijie Meng\textsuperscript{4},
  Teng Xi\textsuperscript{2},
  Yuzhuo Chen\textsuperscript{5},
  Yucheng Wang\textsuperscript{3},\\
  Ruicheng Zhang\textsuperscript{1},
  Guoxin Zhang\textsuperscript{2},
  Pengfei Wan\textsuperscript{2},
  Xiu Li\textsuperscript{1},
  Fangneng Zhan\textsuperscript{3}
  \\[1mm]
  \textsuperscript{1} Shenzhen International Graduate School, Tsinghua University \quad
  \textsuperscript{2} Kling Team \\
  \textsuperscript{3} The Hong Kong University of Science and Technology \quad
  \textsuperscript{4} Peking University \\
  \textsuperscript{5} University of Science and Technology of China  \\
  \textbf{Project: \href{https://liveact-ai.github.io}{\texttt{\textcolor{cyan}{https://liveact-ai.github.io}}}}
}

\begin{document}

\twocolumn[{
\begin{center}
\maketitle
\vspace{-2em}
    \captionsetup{type=figure}
    \includegraphics[width=1.0\textwidth]{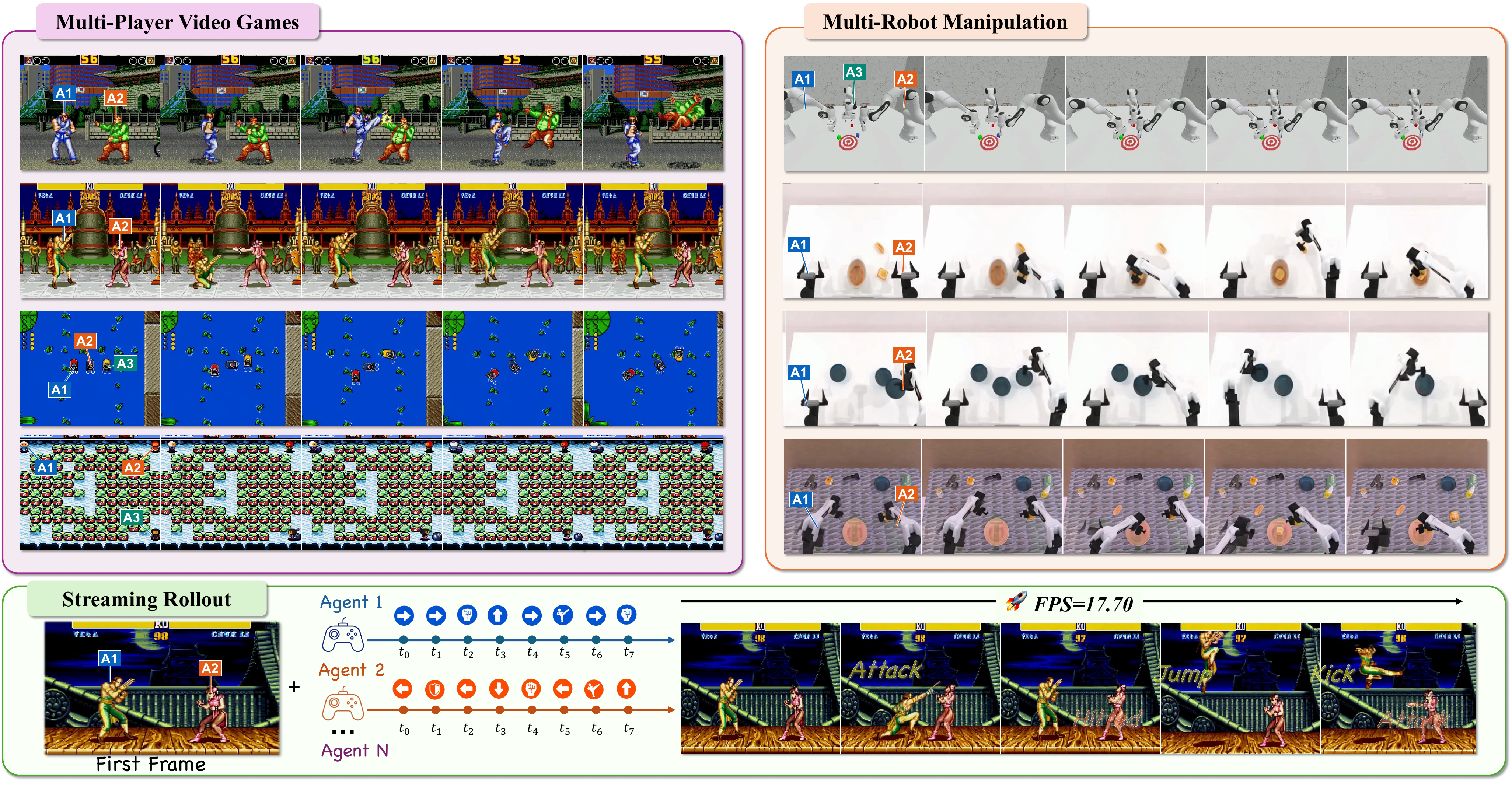}
    \vspace{-1.5em}
    \caption{\textbf{Agent-specific control across heterogeneous domains.} Given an initial frame and synchronized action streams, \ours{} generates shared visual futures for multi-agent games (top left) and multi-robot manipulation (top right) while preserving stream-to-agent correspondence (bottom).}
    \label{teaser}
    \vspace{1em}
\end{center}
}]

\input{sec/0_abstract}
\input{sec/1_introduction}
\input{sec/2_related}
\input{sec/3_method}
\input{sec/4_experiment}
\input{sec/5_conclusion}

\input{sec/X_suppl}
\clearpage  

{
    \small
    \bibliographystyle{ieeenat_fullname}
    \bibliography{main}
}


\end{document}

%% file: math_commands.tex
\usepackage{amsmath,amsfonts,bm}

\def\eqref#1{equation~\ref{#1}}

\def\1{\bm{1}}

\DeclareMathAlphabet{\mathsfit}{\encodingdefault}{\sfdefault}{m}{sl}
\SetMathAlphabet{\mathsfit}{bold}{\encodingdefault}{\sfdefault}{bx}{n}



%% file: preamble.tex
\usepackage{tabularx}
\usepackage{multirow}
\usepackage{makecell}
\usepackage{float}
\usepackage{titletoc}
\usepackage{algorithm}
\usepackage{algpseudocode}
\usepackage{tikz}
\usetikzlibrary{tikzmark,calc}

\definecolor{Red}{RGB}{192,0,0}
\definecolor{Blue}{RGB}{12,114,186}
\definecolor{linkc}{rgb}{0, 0.44, 0.74}
\definecolor{eqc}{rgb}{1, 0, 0}

\definecolor{LiveStart}{HTML}{355CDE}
\definecolor{LiveEnd}{HTML}{20C997}

\DeclareRobustCommand{\LiveScaleLogo}{%
\textcolor{LiveStart}{L}%
\textcolor{LiveStart!80!LiveEnd}{i}%
\textcolor{LiveStart!65!LiveEnd}{v}%
\textcolor{LiveStart!50!LiveEnd}{e}%
\textcolor{LiveStart!35!LiveEnd}{A}%
\textcolor{LiveStart!20!LiveEnd}{c}%
\textcolor{LiveEnd}{t}%
}

\newcommand{\ours}{{LiveAct}}

\newcommand{\mainref}[1]{\ref{#1}}

%% file: sec/0_abstract.tex
\begin{abstract}

Interactive video world models predict how an environment evolves under user actions, but most existing systems assume one controlled agent and one task-specific action interface. We study multi-agent video world modeling, where one model generates a joint visual future for multiple controlled agents while interpreting heterogeneous controls ranging from binary game buttons to continuous robot commands. This task presents two complementary challenges. First, every agent-indexed action stream must control its intended visual entity independently. Second, the model must adapt the meaning of each control variable across different tasks, where the same tensor dimension may denote an attack, steering command, or robot command. 
We introduce LiveAct, centered on a Heterogeneous Action Interface (HAI) whose two components mirror this decomposition. To support independent control, Agent-Aware Spatial Modulation uses language-driven spatial gates and agent-specific attention to bind each action stream to its intended visual agent. To accommodate heterogeneous controls, Schema-Guided Action Adaptation describes each control variable in language and modulates its semantic token with the synchronized numerical value. We further transfer the complete interface to a blockwise causal generator through teacher forcing and rollout-aware distribution matching, enabling efficient streaming generation while retaining agent-specific control and action semantics.
 We instantiate the formulation on six multi-agent sources spanning games and multi-robot manipulation, and evaluate visual quality, action following, agent attribution, and latency. Across games and robot manipulation, LiveAct reaches 0.898 and 0.942 agent-attribution accuracy without agent locations or identity crops, and the causal generator attains 17.70 FPS at $512\times 448$. 
\end{abstract}

%% file: sec/1_introduction.tex
\section{Introduction}
\label{sec:introduction}

Video world models~\citep{team2025hunyuanworld, zhang2026echowm, hong2025relic, ma2024followpose, bi2026motus, li2025hunyuan, zhang2026robostereo, zhang2025mind, hu2026multiplayer, liu2026gamma, king2026echo} turn interaction into conditional generation~\cite{ma2025controllable, he2025magicman, he2024co, zhang2026zo3t, zhang2026kvpo}: given visual history and actions, they synthesize the observations that follow. Such models can provide scalable experience for game simulation, robot learning, planning, and interactive content creation~\citep{hafner2023mastering,yang2023learning}. Existing systems increasingly support persistent and low-latency rollout, but most still expose a single aggregate player or camera-control stream~\citep{valevski2025diffusion,gao2026infinite}. Such a generation paradigm is insufficient for inherently multi-agent environments, where independently controlled entities act concurrently and jointly determine the visual future. We therefore study \emph{multi-agent video world modeling}: given an initial frame, a scene description, and a set of frame-level action streams for multiple agents, the model generates a shared visual future in which each stream controls its intended agent while all agents remain consistent with one another and the environment.

\begin{figure}[t]
  \centering
  \includegraphics[width=\linewidth]{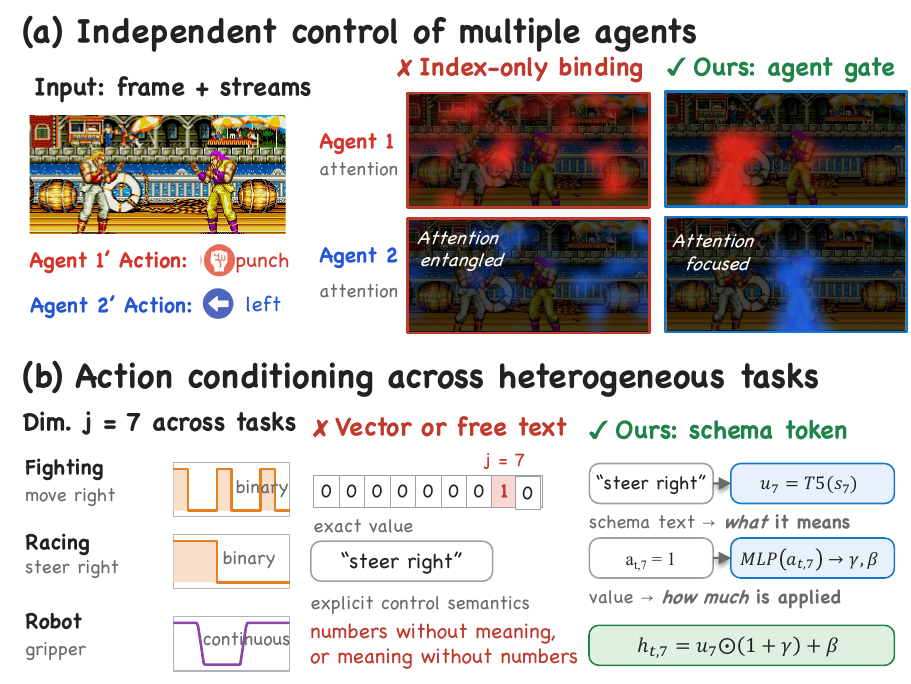}
  \vspace{-1.5em}
  \caption{\textbf{Two challenges for multi-agent action interfaces.} (a) Streams that differ only by input order entangle fighters under index-only binding; \ours{}'s agent gate keeps each on its fighter. (b) Dimension $j{=}7$ changes meaning across tasks. A raw vector keeps the value but not the meaning, free text the reverse; \ours{} modulates each schema token by its value.}
  \label{fig:challenges}
  \vspace{-1em}
\end{figure}

Existing interactive world models cover this setting only in part, and the shortfall lies in the action interface rather than the video backbone. Multi-agent control has relied on numerical action vectors, which keep exact values but typically reach the right entity through spatial priors supplied at inference time and leave control semantics implicit in task-specific dimensions~\citep{wu2026multiworld,pondaven2026actionparty,hu2026wantofight}. Language-based actions transfer semantics across entities and tasks but give up the synchronized numerical values that continuous control requires~\citep{zhu2026incantation}. Tab.~\ref{tab:capability_comparison} summarizes the landscape: no current interface is at once multi-agent, real-time, semantically grounded, and numerically exact. We ask whether one interface can satisfy all four requirements without inference-time masks, boxes, poses, or reference images.

However, a unified action interface must still address two challenges: \emph{who} each action stream should control in a shared visual scene, and \emph{what} each control dimension means across heterogeneous tasks. These requirements lead to two key challenges.

\begin{enumerate}[label=(\arabic*), nosep, leftmargin=*]

\item  \textit{\textbf{maintaining independent control over multiple interacting agents.}} A model must keep each agent responsive to its own action stream while capturing interactions. The action tensors of each agent are injected sequentially at input, but do not ensure that their effects remain correctly attributed in the shared visual feature space. Commands may be assigned to the wrong agent or inadvertently drive multiple entities (Fig.~\ref{fig:challenges}a), especially when agents overlap, look similar, or jointly manipulate an object. Achieving independent control therefore requires agent-specific conditioning that separates command attribution.

\item  \textit{\textbf{unifying action conditioning across heterogeneous control spaces.}} Action vectors do not share a universal semantic coordinate system: their dimensions, ordering, and meanings are defined by task-specific action spaces. As a result, the same action dimension or pattern may encode different behaviors (Fig.~\ref{fig:challenges}b). Directly conditioning on raw vectors therefore leads to action semantics being tied to task-specific dimensions and ordering. Language can expose what each control dimension means, but does not preserve the exact frame-synchronized binary or continuous values needed for execution. Unified action conditioning thus requires separating \emph{what a control means} from \emph{what value it takes}, and representing both in a shared action interface.

\end{enumerate}

\input{tab/capability_comparison}

\ours{} addresses these challenges with a \emph{Heterogeneous Action Interface} (HAI), which factorizes action conditioning into semantic adaptation and agent-specific visual injection. For \emph{independent control}, \emph{Agent-Aware Spatial Modulation} derives an agent-specific spatial gate from the agent description, so that each branch selectively updates the visual tokens associated with its intended entity while attending only to the corresponding action stream. For \emph{heterogeneous action conditioning}, \emph{Schema-Guided Action Adaptation} converts each control dimension into a semantic token defined by its schema and injects the synchronized numerical value through affine modulation, decoupling control meaning from task-specific vector layouts. This provides a shared semantic interface for heterogeneous control spaces. Together, these two designs provide a common action representation while preserving agent-specific effects in the shared visual space. The resulting residuals are injected through zero-initialized projections, preserving the pretrained visual prior at initialization.

Beyond accurate control, interaction requires efficient incremental rollout without repeatedly regenerating the visual history. We therefore retain the complete HAI when transferring a bidirectional video teacher to a blockwise causal generator through Teacher Forcing for autoregressive (AR) Tuning and Distribution Matching Distillation (DMD) for Streaming Video~\citep{yin2024one,huang2026self, wang2026subjectanchor, wang2026liveedit}. Causal temporal alignment maps raw actions to compressed video latents, while a global latent offset advances action windows together with the cached rollout. Causal generation thus accelerates a fixed control interface rather than redefining who acts or what each action means.

We evaluate \ours{} across six multi-agent environments spanning interactive games and robot manipulation, covering both discrete and continuous control spaces. Experiments assess visual fidelity, action adherence, agent attribution, and real-time efficiency. Our main contributions are:

\begin{itemize}[leftmargin=*,nosep]
    \item We formulate real-time multi-agent video world modeling under heterogeneous action spaces and identify two coupled challenges overlooked by existing action interfaces: binding each indexed action stream to its intended visual agent and jointly representing task-specific control semantics and synchronized numerical values in a shared interface.
    \item We propose \ours{}, a real-time multi-agent video world model built around a Heterogeneous Action Interface. HAI combines Agent-Aware Spatial Modulation for language-driven stream-to-agent binding with Schema-Guided Action Adaptation for value-preserving semantic control, and is retained during teacher-forced causal adaptation and rollout-aware distribution matching for real-time streaming generation.
    \item We build a dataset of 17,951 multi-agent trajectories, totaling 119.54 hours of raw video, and a 108-clip benchmark spanning four games and two robot-manipulation sources. \ours{} reaches 0.898 and 0.942 agent-attribution accuracy on games and robots without locations, masks, or identity crops, and generates at 17.70 FPS at $512\times448$.
\end{itemize}

%% file: tab/capability_comparison.tex
\begin{table*}[t]
\centering
    \vspace{-0.5em}
    \caption{\textbf{Native capability comparison.} $\checkmark$: supported; $\times$: unsupported.}
    \vspace{-1em}
    \label{tab:capability_comparison}
    \resizebox{0.98\textwidth}{!}{%
        \begin{tabular}{llcccc}
        \toprule
        \textbf{Method} & \textbf{Domain} & \textbf{Multi-Agent} & \textbf{Real-Time} & \textbf{Semantic} & \textbf{Numerical} \\
        \midrule
        LingBot-World~2.0~\citep{gao2026infinite}& General visual worlds                  & $\times$     & $\checkmark$ & $\checkmark$ & $\checkmark$ \\
        Solaris~\citep{savva2026solaris}           & Single (\textit{Minecraft})               & $\checkmark$ & $\times$     & $\times$     & $\checkmark$ \\
        MultiWorld~\citep{wu2026multiworld}         & Single (\textit{It Takes Two})           & $\checkmark$ & $\times$     & $\times$     & $\checkmark$ \\
        ActionParty~\citep{pondaven2026actionparty}& Single (\textit{Melting Pot})             & $\checkmark$ & $\times$     & $\times$     & $\checkmark$ \\
        Incantation~\citep{zhu2026incantation}     & Single (\textit{Elden Ring})          & $\checkmark$ & $\checkmark$ & $\checkmark$ & $\times$     \\
        WanToFight~\citep{hu2026wantofight}        & Single (KOF~'97)                          & $\checkmark$ & $\checkmark$ & $\times$     & $\checkmark$ \\
        \midrule
        \textbf{\ours{} (Ours)}                    & \textbf{Diverse} & $\checkmark$ & $\checkmark$ & $\checkmark$ & $\checkmark$ \\
        \bottomrule
        \end{tabular}%
    }
\vspace{-1em}
\end{table*}

%% file: sec/2_related.tex
\section{Related Work}
\label{sec:related}

\noindent\textbf{Multi-agent video world models.}
Multi-agent video world models jointly predict an environment under multiple agents' observations or controls, and must coordinate shared dynamics while preserving the identity of each condition stream. Solaris jointly models synchronized observations from multiple Minecraft agents~\citep{savva2026solaris}; MultiWorld combines a multi-agent condition module with a global state encoder across game and robot manipulation~\citep{wu2026multiworld}; and Gamma-World uses permutation-equivalent agent encoding and sparse hub attention~\citep{liu2026gamma}. Complementary systems focus on binding independent controls to visual entities: ActionParty jointly predicts persistent subject-state tokens initialized from subject locations~\citep{pondaven2026actionparty}, WanToFight grounds locally causal keyboard streams with first-frame character references~\citep{hu2026wantofight}, and Incantation binds language actions to individual entities~\citep{zhu2026incantation}. Across these designs, the central control problem is to preserve per-agent condition provenance and ensure that each action affects its intended entity. The Agent-Aware Spatial Modulation component of our HAI addresses this problem by conditioning the visual queries on each agent's language description. without masks, boxes, coordinates, or reference images at inference time.

\noindent\textbf{Action-controllable world models across domains.}
World models learn environment dynamics for imagined interaction~\citep{hafner2023mastering}. Genie learns latent actions from unlabeled videos~\citep{bruce2024genie}, DIAMOND uses diffusion for Atari and Counter-Strike dynamics~\citep{alonso2024diffusion}, and GameNGen predicts DOOM frames from past observations and actions~\citep{valevski2025diffusion}. LingBot-World~2.0 couples camera poses and time-varying semantic prompts for open-ended interaction~\citep{gao2026infinite}. Real-world simulators extend action-conditioned prediction to diverse interactions~\citep{yang2023learning} and continuous robot trajectories~\citep{zhu2025irasim}, while Incantation expresses per-entity actions in natural language to support semantic transfer~\citep{zhu2026incantation}. 
Most interfaces choose between numerically precise but task-specific vectors and language interfaces that make control semantics explicit.
Within our HAI, Schema-Guided Action Adaptation combines the two: language identifies the task-dependent meaning of each control variable, while synchronized binary or continuous values retain frame-level control.

%% file: sec/3_method.tex
\section{Method}
\label{sec:method}
\begin{figure*}[t]
    \centering
    \includegraphics[width=0.98\textwidth]{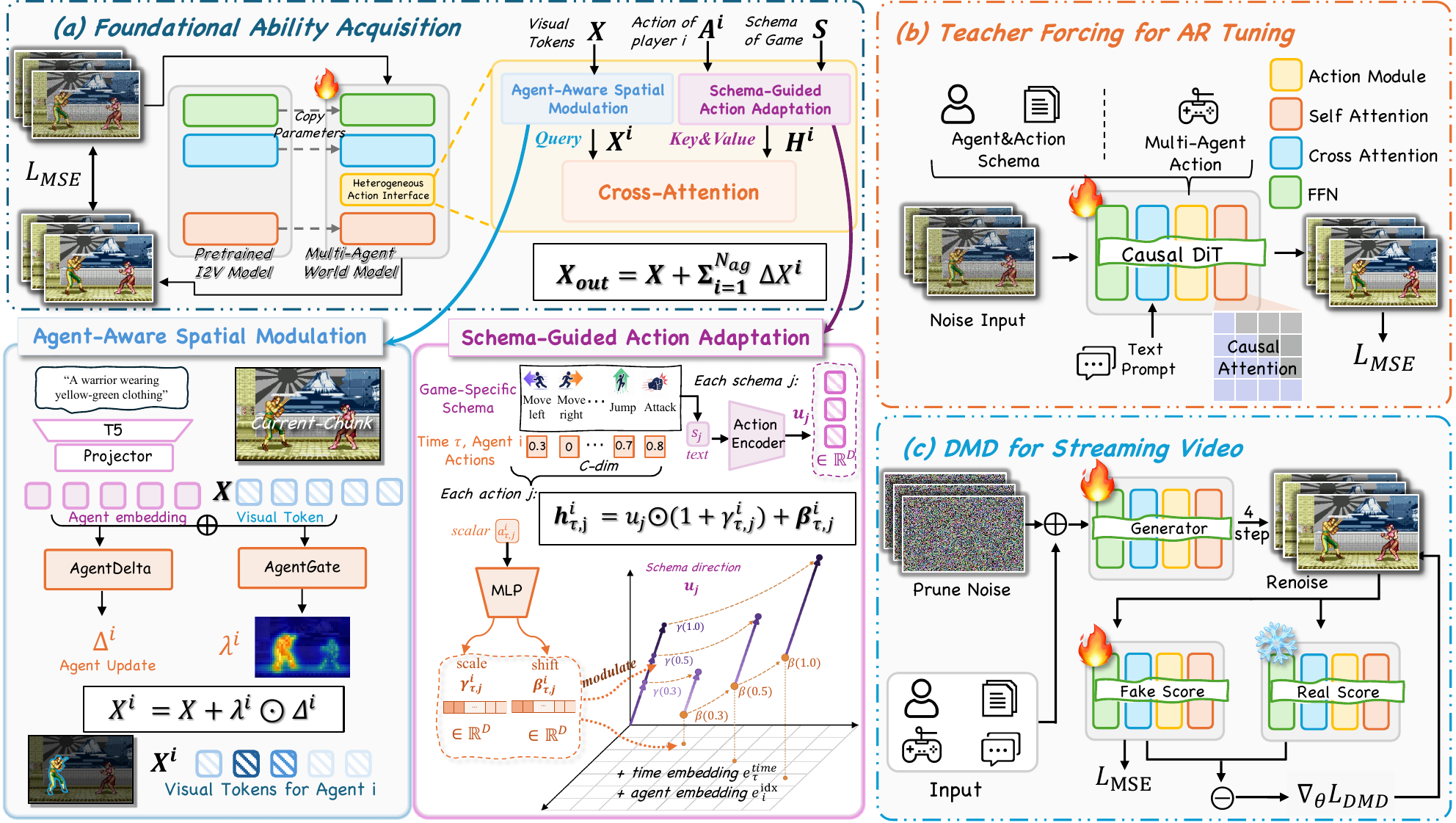}
    \caption{\textbf{Overview of \ours.} (a) Foundational Ability Acquisition trains a bidirectional teacher with a Heterogeneous Action Interface, which integrates Agent-Aware Spatial Modulation and Schema-Guided Action Adaptation. (b) Teacher Forcing for AR Tuning transfers the same HAI to causal attention. (c) DMD for Streaming Video distills few-step streaming rollouts.}
    \label{fig:overview}
    \vspace{-1.5em}
\end{figure*}

Given an initial frame $I_0$, a scene description $c$, and ordered action streams $\mathcal{A}=\{A^{(i)}\}_{i=1}^{N_{\mathrm{ag}}}$ for ${N_{\mathrm{ag}}}$ agents, where $A^{(i)}=[a_{\tau,j}^{(i)}]\in\mathbb{R}^{T\times C}$ gives the value of control dimension $j$ at raw frame $\tau\in\{1,\ldots,T\}$ (raw frame $0$ is $I_0$), paired with agent descriptions $G=\{g_i\}$ and a control schema $S=\{s_j\}_{j=1}^{C}$, multi-agent world modeling aims to synthesize the shared visual future in which each stream drives its own agent while all agents remain consistent with one another and with the scene. We write $\mathcal{C}=(c,\mathcal{A},S,G)$ for the full condition. In this section, we first discuss the motivation and two failure modes of existing action interfaces in Sec.~\ref{sec:motivation}. We then introduce Agent-Aware Spatial Modulation in Sec.~\ref{sec:agent_control}, which decides \emph{who} responds to each stream, and Schema-Guided Action Adaptation in Sec.~\ref{sec:action_conditioner}, \emph{what} each control dimension means and \emph{what value} it takes; Sec.~\ref{sec:action_module_integration} assembles them into the Heterogeneous Action Interface. Finally, Sec.~\ref{sec:efficient_rollout} carries the same action-conditioning mechanism into causal streaming rollout. Fig.~\ref{fig:overview} summarizes this three-stage design.

\subsection{Motivation}
\label{sec:motivation}

Although the indexed action tensor preserves the numerical trajectory of each stream, it does not establish the correspondences required to condition a shared visual representation. The stream index $i$ separates $A^{(i)}$ from the other streams but provides no visual evidence associating it with the tokens of its intended agent. Likewise, the control index $j$ identifies a coordinate only within a task-specific action space and therefore lacks a consistent interpretation across tasks. 

\noindent\textbf{Cross-Agent Misattribution.} Keeping the action streams separate preserves their provenance in action space, but does not establish their recipients in visual space. An agent-index embedding can distinguish $A^{(i)}$ from $A^{(i')}$, yet contains no visual evidence linking index $i$ to the appearance of agent $i$. Consequently, the command from one stream may be attributed to the wrong visual agent or inadvertently affect multiple agents. The missing correspondence must therefore be established on the query side of action conditioning: the visual features that read $A^{(i)}$ should be grounded in the description $g_i$ of its corresponding agent.

\noindent\textbf{Control-Semantics Collapse.} An action vector contains $C$ control dimensions, where the $j$-th coordinate provides the numerical value of one control at a given time step. However, the index $j$ itself carries no explicit semantic meaning beyond the definition of its task-specific action space. Compressing all $C$ dimensions into a single action embedding further obscures the identity of individual controls, making it difficult for the model to distinguish which control produces a particular effect. The action representation should therefore keep each control dimension distinct while jointly encoding its schema-defined semantics and frame-level numerical value.

These two requirements are complementary: expressive action tokens cannot resolve agent binding, while agent-aware queries cannot recover missing control semantics. HAI therefore separates the two, using Agent-Aware Spatial Modulation for query-side binding and Schema-Guided Action Adaptation for key--value-side control representation, coupled through cross-attention.

\subsection{Agent-Aware Spatial Modulation}
\label{sec:agent_control}

To prevent cross-agent misattribution, we condition the visual queries on each agent's description $g_i$, encouraging each action stream to affect its corresponding visual agent without requiring explicit spatial supervision or inference-time spatial inputs.

\paragraph{Agent semantic representations.}

By construction, each action sequence $A^{(i)}$ is paired with an agent description $g_i$. We encode this description with the frozen umT5~\citep{2020t5} text encoder and project it to an agent embedding:
\begin{equation}
e_i^{G}=\operatorname{Pool}(\operatorname{T5}(g_i))\in\mathbb{R}^{4096},\quad
q_i=\operatorname{Proj}_{G}(e_i^G)\in\mathbb{R}^{D}.
\label{eq:agent_encoding}
\end{equation}
Here $\operatorname{Pool}$ denotes a pooling layer and $D$ is the backbone width. The resulting $q_i$ provides the semantic identity associated with action stream $A^{(i)}$.

\paragraph{Spatial query modulation.}

A global identity vector does not by itself indicate where the corresponding agent appears. For latent frame $k$, let $X_k=[x_{k,1},\ldots,x_{k,N}]^\top\in\mathbb{R}^{N\times D}$ contain the $N$ patchified visual tokens. We compute a token-dependent gate from the visual feature, the agent representation, and their element-wise interaction:

\begin{equation}
\begin{aligned}
\lambda_{k,n}^{(i)}&=\operatorname{sigmoid}\!\left(\operatorname{AgentGate}
([x_{k,n},q_i,x_{k,n}\odot q_i])\right),\\
x_{k,n}^{(i)}&=x_{k,n}+\lambda_{k,n}^{(i)}\operatorname{AgentDelta}(q_i).
\end{aligned}
\label{eq:agent_residual}
\end{equation}
Here, $\operatorname{AgentDelta}(q_i)\in\mathbb{R}^{D}$ supplies an identity-dependent feature direction, while the scalar $\lambda_{k,n}^{(i)}$ controls its strength at each spatial token. Because $x_{k,n}$ varies across space and time, the same agent description can induce different modulation strengths throughout the video. The gate is a soft, learned modulation rather than a segmentation mask or an explicitly supervised object assignment.

Stacking the modulated tokens gives $X_k^{(i)}=[x_{k,1}^{(i)},\ldots,x_{k,N}^{(i)}]^\top$. Each $X_k^{(i)}$ is computed from the same unmodified state $X_k$ and only preconditions the queries for agent $i$; it is not written into the backbone as a separate residual. The agent-conditioned states remain separate until cross-attention forms the action residuals.

\subsection{Schema-Guided Action Adaptation}
\label{sec:action_conditioner}

To prevent control-semantics collapse, we construct a separate action token for each control dimension $j$ at each time step $\tau$. Each token combines the schema description $s_j$, which specifies \emph{what} the control means, with its synchronized numerical value $a_{\tau,j}^{(i)}$, which specifies its current state. The resulting tokens are used as keys and values in cross-attention.

\paragraph{Semantic control representations.}

Raw dimension indices cannot express task-specific control semantics. We therefore encode description $s_j$ of every control variable with the same text encoder:
\begin{equation}
e_j^{S}=\operatorname{Pool}(\operatorname{T5}(s_j))\in\mathbb{R}^{4096},
\quad u_j=\operatorname{Proj}_{S}(e_j^S)\in\mathbb{R}^{D}.
\label{eq:schema_encoding}
\end{equation}
The resulting token turns coordinate $j$ into a named control such as an attack, steering command, or robot command. The schema description and numerical value are paired by control dimension, allowing a shared representation to accommodate heterogeneous control layouts within a single model.

\paragraph{Value-modulated control tokens.}

Language alone expresses action meaning but does not preserve the synchronized binary or continuous trajectory. For agent $i$, raw-frame index $\tau$, and control dimension $j$, a value encoder shared across control dimensions maps the scalar $a_{\tau,j}^{(i)}$ to affine vectors:
\begin{equation}
\begin{aligned}
(\gamma_{\tau,j}^{(i)},\beta_{\tau,j}^{(i)})
&=\operatorname{MLP}_{\mathrm{mod}}(a_{\tau,j}^{(i)}),\\
h_{\tau,j}^{(i)}&=u_j\odot(1+\gamma_{\tau,j}^{(i)})+\beta_{\tau,j}^{(i)}.
\end{aligned}
\label{eq:modulated_control}
\end{equation}
Thus, $u_j$ specifies \emph{what} a control variable represents, while the synchronized value determines its current modulation. Unlike early aggregation, this construction retains $C$ tokens per time step, allowing visual queries to select individual controls. We add learned time and agent embeddings,
\begin{equation}
\widetilde h_{\tau,j}^{(i)}=h_{\tau,j}^{(i)}+e_\tau^{\mathrm{time}}+e_i^{\mathrm{idx}},
\label{eq:action_token}
\end{equation}
and collect these tokens into $\widetilde H^{(i)}\in\mathbb{R}^{T\times C\times D}$, retaining a separate token for every control dimension at every time step before temporal windowing.

\subsection{Assembling the Heterogeneous Action Interface}
\label{sec:action_module_integration}

After constructing the agent-conditioned visual queries and schema-guided action tokens, we assemble the HAI through per-agent cross-attention. For latent frame $k$ and agent $i$, $X_k^{(i)}\in\mathbb{R}^{N\times D}$ queries the aligned action window $H_k^{(i)}\in\mathbb{R}^{M\times D}$ (Eq.~\ref{eq:causal_action_memory}), where $r$ is the temporal compression ratio of the video VAE and $M=rC$ in the one-latent window used by our main model:
\begin{equation}
\Delta X_k^{(i)}=\operatorname{CrossAttn}(X_k^{(i)},H_k^{(i)}).
\label{eq:player_action_attention}
\end{equation}
The visual tokens serve as queries and the action tokens as keys and values, producing one control residual per agent. We insert this layer after text cross-attention and before the feed-forward sublayer in each transformer block. The shared visual state $X_k$ of latent frame $k$ is then updated as
\begin{equation}
\widehat X_k= X_k
+\sum\nolimits_{i=1}^{N_{\mathrm{ag}}}\Delta X_k^{(i)}.
\label{eq:block_action_injection}
\end{equation}
All agents read from the same shared visual state $X_k$, while their control residuals are computed independently before aggregation. We zero-initialize each cross-attention output projection, so that the augmented model initially preserves the pretrained backbone behavior.

\subsection{Efficient Causal Rollout}
\label{sec:efficient_rollout}

To enable streaming rollouts, we convert the bidirectional diffusion teacher into a causal generator while addressing action alignment and exposure to self-generated history. Teacher Forcing for AR Tuning first adapts the model to causal attention, followed by DMD on autoregressive rollouts.

\paragraph{Causally aligned action conditioning.}

Each video latent should read only the controls of its own physical interval. Since the VAE encodes $I_0$ separately and compresses every $r$ raw frames into one latent, future latent $k\geq1$ covers raw frames $\mathcal{I}_k=\{r(k-1)+1,\ldots,rk\}$, and the action memory of agent $i$ stacks the control tokens in this interval:
\begin{equation}
H_k^{(i)}=\big[\widetilde h_{\tau,j}^{(i)}\big]_{\tau\in\mathcal{I}_k,\;1\le j\le C}\in\mathbb{R}^{rC\times D}.
\label{eq:causal_action_memory}
\end{equation}
In the causal generator, $k$ is the global latent index rather than the position within the current chunk, so every latent keeps its physical controls as the KV cache advances.

\paragraph{Teacher Forcing for AR Tuning.}

After obtaining a bidirectional model, this stage then transfers its parameters to a blockwise causal generator~\citep{zhu2026causal}. Each target chunk is denoised using only clean preceding chunks, with a causal mask. The HAI remains at the same transformer locations and receives the same aligned conditions, so this stage isolates the change in temporal information flow from errors caused by self-generated history. At inference, preceding chunks are retained in the KV cache and reused when generating the next one.

\paragraph{DMD for Streaming Video.}

Teacher forcing still trains on clean history, leaving an exposure gap at deployment. DMD addresses this gap by optimizing autoregressive rollouts~\citep{yin2024one,huang2026self}. Starting from the initial latent $z_0$, the VAE encoding of $I_0$, the generator produces $\hat z=\mathcal{G}_\theta(z_0,\mathcal{C})$ while reusing its self-generated KV cache. The frozen bidirectional teacher serves as the real score, while a trainable bidirectional fake score initialized from the same teacher tracks the generator distribution. At diffusion timestep $t$, their denoised predictions from the same noised sample $\hat z_t$ and condition $\mathcal{C}$ define the unnormalized distribution-matching gradient
$\delta_{\mathrm{DMD}}=
\mu_{\mathrm{fake},\phi}(\hat z_t,t,\mathcal{C})
-\mu_{\mathrm{real}}(\hat z_t,t,\mathcal{C}).$
Backpropagating the normalized gradient through $\mathcal{G}_\theta$ exposes the causal model to its own rollout history while retaining few-step chunk-wise generation.

%% file: sec/4_experiment.tex
\section{Experiments}
\label{sec:experiments}

\begin{figure*}[t]
    \centering
    \includegraphics[width=0.98\textwidth]{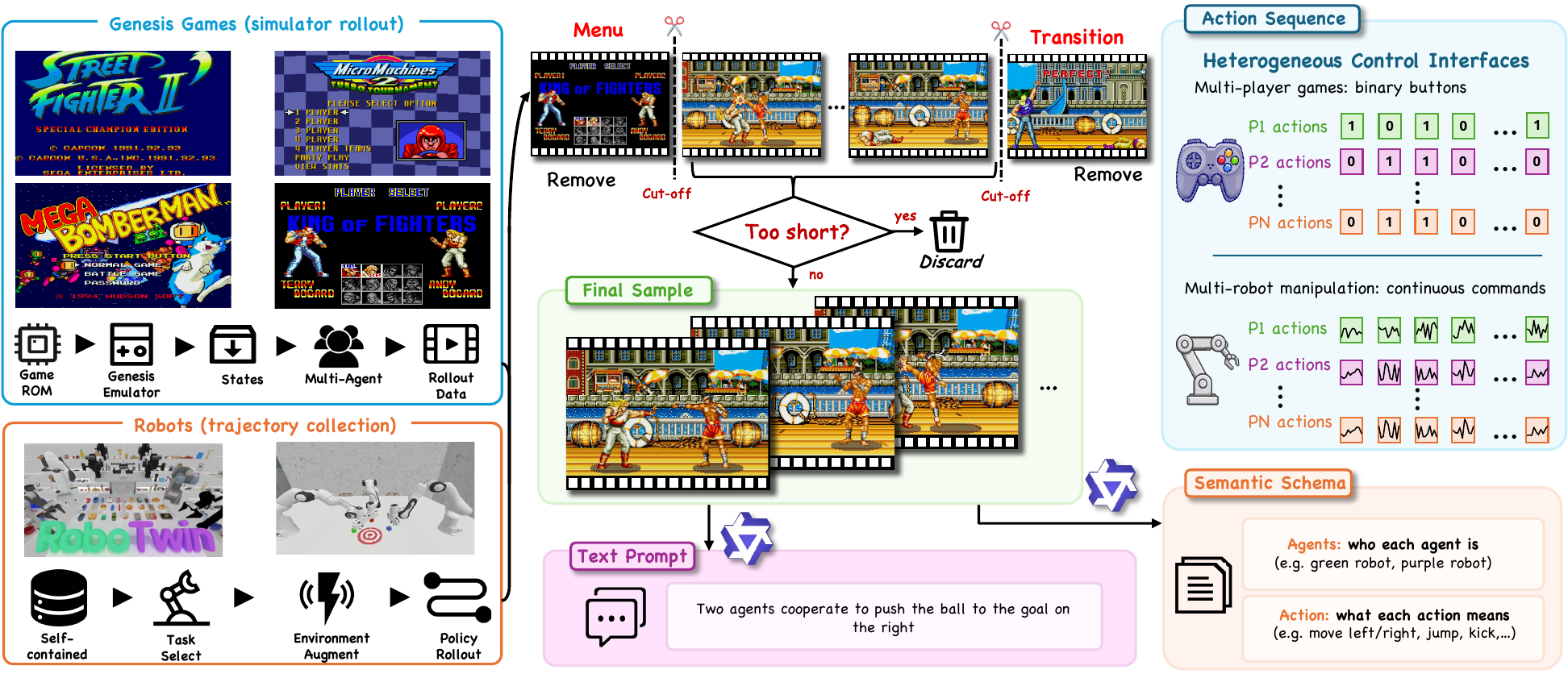}
    \caption{\textbf{Data curation pipeline.} Game and robotic episodes are curated into aligned samples containing a scene prompt,  per-agent action streams, agent descriptions, and task-specific schemas.}
    \label{fig:data_pipeline}
    \vspace{-1em}
\end{figure*}

\subsection{Data Collection}
\label{sec:data}

As shown in Fig.~\ref{fig:data_pipeline}, we assemble a six-source dataset of 17,951 independent multi-agent trajectories from four competitive multiplayer games and collaborative multi-robotic sources, totaling 119.54 hours of raw video. Each sample contains an initial frame, a scene description, synchronized per-agent action streams, agent descriptions, and a task-specific control schema, with binary controls for games and continuous controls for robots. For evaluation, we build a benchmark of 108 clips drawn from trajectories held out from training, with one 81-frame clip per trajectory; all comparisons and ablations in this section are conducted on this benchmark. Details are provided in Appendix.

\subsection{Implementation details}

We build \ours{} on Wan2.2-TI2V-5B~\citep{wan2025wan}, which contains 30 transformer blocks with a hidden width of $D=3072$ and 24 attention heads; its VAE compresses every $r=4$ raw frames into one latent. We insert the HAI after text cross-attention and before the feed-forward sublayer in every block and fine-tune the full DiT together with the HAI, while the umT5 encoder, which encodes agent and schema descriptions, and the VAE remain frozen. Each agent is represented by $C=12$ schema-aligned control dimensions. Stage 1 trains the bidirectional teacher for 10k steps with a learning rate of $10^{-4}$ and a global batch size of 16. Stage 2 then trains the causal generator for 6k steps with a learning rate of $2\times10^{-6}$, and Stage 3 runs 700 iterations with learning rates of $2\times10^{-6}$ for the generator and $4\times10^{-7}$ for the fake score; both causal stages use a global batch size of 24 with FSDP. At inference, the teacher uses 50 denoising steps, and the causal generator predicts four latent frames per chunk with four denoising evaluations at timesteps $[1000,750,500,250]$.

\subsection{Comparison with Baselines}
\label{sec:main_results}

\begin{figure*}[t]
    \centering
    \includegraphics[width=1.0\textwidth]{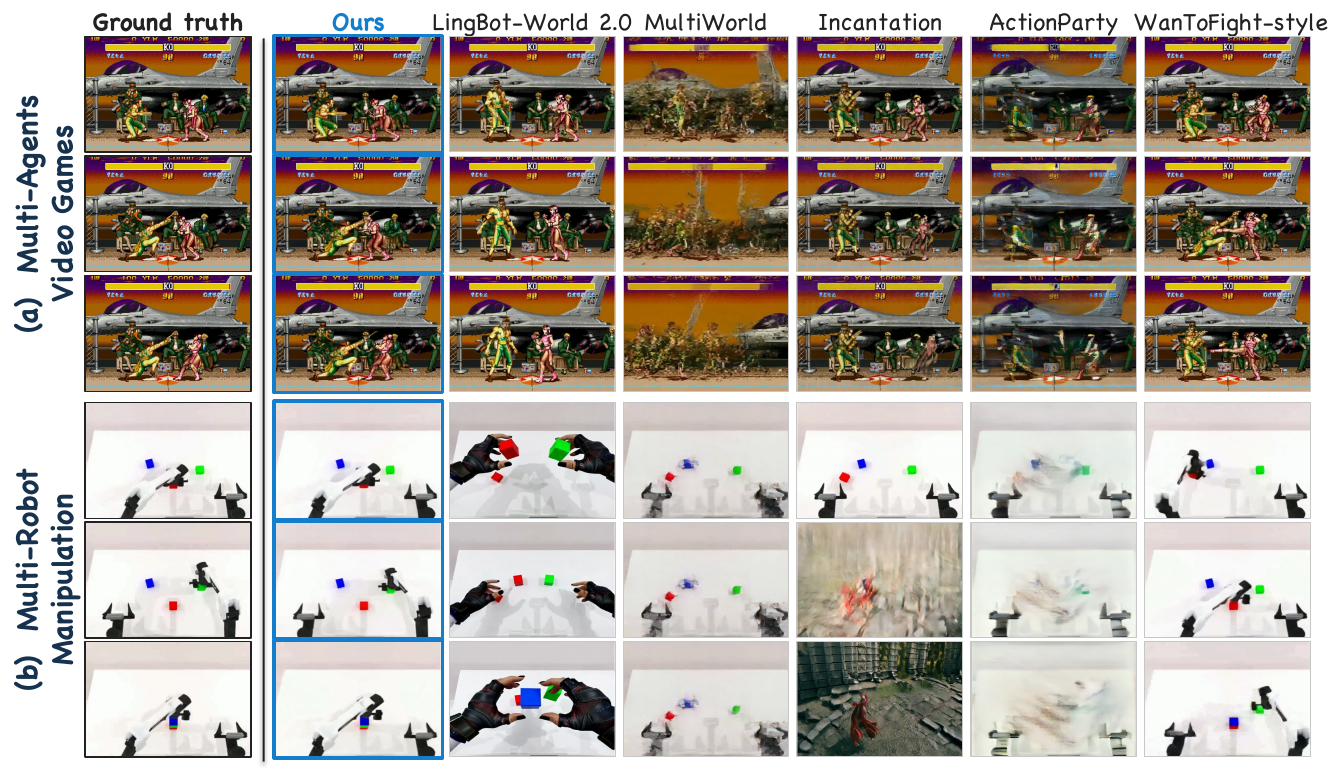}
    \vspace{-1em}
    \caption{\textbf{Qualitative comparison on shared initial frames and action streams.} Columns show ground truth, the five baselines, and \ours; Rows are frames sampled at different durations.}
    \label{fig:qualitative}
\end{figure*}

\noindent\textbf{Baselines.}
We evaluate \ours{} on two complementary tracks---\emph{Multi-Agent Video Games} and \emph{Multi-Robot Manipulation}---under the same protocol for visual quality, action following, and agent attribution. We compare against representative interactive and multi-agent video world models covering three control paradigms: language-serialized actions (LingBot-World~2.0~\citep{gao2026infinite}, Incantation~\citep{zhu2026incantation}), per-agent numerical controls (MultiWorld~\citep{wu2026multiworld}, ActionParty~\citep{pondaven2026actionparty}), and reference-guided method (WanToFight-style~\citep{hu2026wantofight}).

\noindent\textbf{Metrics.}
Following MultiWorld, we report FVD, LPIPS, PSNR, and SSIM for reconstruction quality. Because frame-paired metrics can favor static or averaged predictions, we complement them with Motion Magnitude Error (MME) and an Aesthetic score using the LAION~\citep{schuhmann2022laion} improved predictor. For a more comprehensive control evaluation, we trained inverse dynamics models only on real clips: they recover the executed controls from a generated video, yielding Act-F1 over discrete game events and a normalized error (NRMSE) on continuous robot commands. To evaluate agent attribution, Attr-Acc measures whether the recovered per-agent controls match the commanded streams better under the intended agent assignment than under pairwise agent swaps; higher is better, with a chance level of 0.5. See appendix for more details.

\noindent\textbf{Quantitative Results.}
Shown in Tab.~\ref{tab:main}, among the four non-reference-conditioned methods, LingBot-World~2.0 gives the best game FVD and Aesthetic score but follows actions weakly; MultiWorld retains competitive robot PSNR and SSIM but has low agent attribution; Incantation reaches comparable game Act-F1 yet suffers severe visual drift on robots; and ActionParty provides the best robot reconstruction but the weakest action following. WanToFight-style is the strongest baseline overall, combining strong reconstruction with 0.886 robot Attr-Acc using identity crops, although its game Attr-Acc remains 0.552. \ours{} obtains the best results among the evaluated methods on both tracks, with FVDs of 338.5 and 270.2, game Act-F1 of 0.594, robot NRMSE of 0.558, and Attr-Acc of 0.898 and 0.942. It also achieves the best MME and obtains the highest Aesthetic score.
 
\input{tab/main_comparison}

\noindent\textbf{Qualitative Results.}
As shown in Fig.~\ref{fig:qualitative}, the language-serialized models lose the robot scene (LingBot-World~2.0, Incantation). On the game track, they keep the stage but not the commanded attacks. MultiWorld and ActionParty keep the coarse layout but blur the agents: the fighters and arms blur into faint, duplicated silhouettes. WanToFight-style preserves both scenes and agents and executes the kick, but misses the second agent's command. \ours{} reproduces the kick and the opponent's response at the same frames as the ground truth and carries the arm along the ground-truth path, showing that each stream is both executed and bound to the intended agent.

\subsection{Ablation Studies}
\label{sec:ablations}

\begin{figure}[t]
  \centering
  \includegraphics[width=\linewidth]{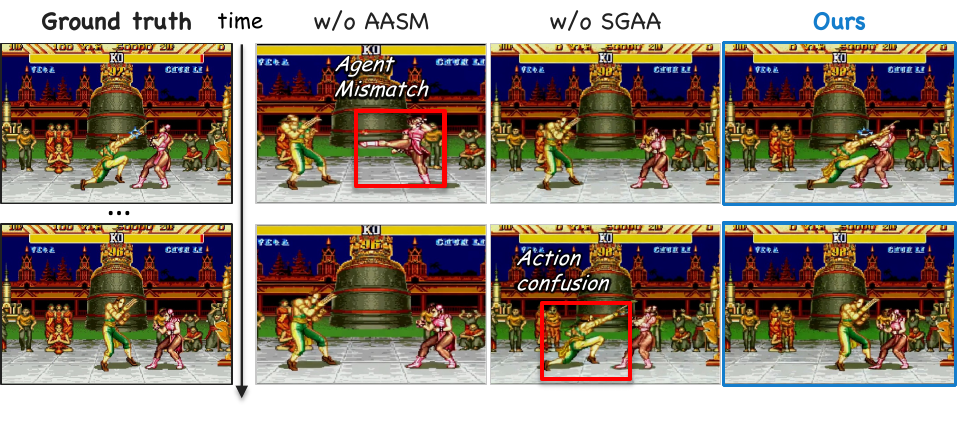}
  \vspace{-1.5em}
  \caption{\textbf{Qualitative ablation study.}}
  \label{fig:ablation}
  \vspace{-1em}
\end{figure}

\noindent\textbf{Effectiveness of Agent-Aware Spatial Modulation (AASM).}
 Removing it keeps the action tokens intact but makes the queries identical for both agents, and control collapses in both domains (Tab.~\ref{tab:ablation}): game Act-F1 drops from 0.594 to 0.172 and Attr-Acc from 0.898 to 0.592, while robot NRMSE rises from 0.558 to 0.931. In Fig.~\ref{fig:ablation}, the two streams no longer stay on their own agents, which is the cross-agent misattribution of Sec.~\ref{sec:motivation}: binding cannot be recovered from the action side alone.

\begin{table}[t]
  \centering
  \caption{\textbf{Quantitative ablation study.}}
  \vspace{-0.5em}
  \label{tab:ablation}
  \setlength{\tabcolsep}{3pt}
  \resizebox{\linewidth}{!}{
  \begin{tabular}{lcccc}
  \toprule
  Variant & Game Act-F1 $\uparrow$ & Game Attr-Acc $\uparrow$ & Robot NRMSE $\downarrow$ & Robot Attr-Acc $\uparrow$ \\
  \midrule
  Full \ours & \textbf{0.594} & \textbf{0.898} & \textbf{0.558} & \textbf{0.942} \\
  \quad w/o AASM & 0.172 & 0.592 & 0.931 & 0.810 \\
  \quad w/o SGAA & 0.225 & 0.603 & 1.120 & 0.484 \\
  \bottomrule
  \end{tabular}}
  \vspace{-1em}
\end{table}

\noindent\textbf{Effectiveness of Schema-Guided Action Adaptation (SGAA).}
Replacing it with a conventional flat action encoder, an MLP that maps each frame's full action vector to a single token, keeps agent modulation but reintroduces control-semantics collapse in visual, and performance again drops in both domains: game Act-F1 falls to 0.225 and Attr-Acc to 0.603, while robot NRMSE rises to 1.120 and Attr-Acc falls to 0.484. Attribution degrades even though agent modulation is retained, indicating that the queries can only route controls they can interpret; the two components are therefore complementary. Finer-grained SGAA ablations are provided in the Appendix.

\noindent\textbf{Effectiveness of real-time distillation.}
Stage 1 denoises the full sequence and cannot stream. Stage 2 enables blockwise causal generation but remains slow at 1.66 FPS. Stage 3 reduces each chunk to four denoising steps and reaches 17.70 FPS, $10.7\times$ faster than Stage 2. On games, Act-F1 rises from 0.539 to 0.594 and Attr-Acc from 0.844 to 0.898. Autoregressive generation is better suited to these discrete actions: each frame is conditioned only on past context and the controls aligned to that frame, rather than denoising the full sequence jointly. On robots, NRMSE rises from 0.430 to 0.558 and Attr-Acc falls from 0.969 to 0.942. Per-stage settings and results are in the Appendix.

%% file: tab/main_comparison.tex
\begin{table*}[!t]
\centering
\caption{\textbf{Main comparison across multi-agent video games and multi-robot manipulation.} Best results are bolded.$^{a}$Native MultiWorld uses synchronized multi-view observations; we have set it to a fixed same perspective. $^{b}$The code is not released; this is our reproduction.}
\vspace{-1em}
\label{tab:main}
\setlength{\tabcolsep}{3.5pt}
\resizebox{0.98\textwidth}{!}{
\begin{tabular}{lcccccccc}
\toprule
& \multicolumn{4}{c}{\textit{Reconstruction}} & \multicolumn{2}{c}{\textit{Control}} & \multicolumn{2}{c}{\textit{Diagnostics}} \\
\cmidrule(lr){2-5} \cmidrule(lr){6-7} \cmidrule(lr){8-9}
Method & FVD $\downarrow$ & LPIPS $\downarrow$ & PSNR $\uparrow$ & SSIM $\uparrow$
& Act-F1 $\uparrow$ / NRMSE $\downarrow$ & Attr-Acc $\uparrow$ & MME $\downarrow$ & Aesthetic $\uparrow$ \\
\midrule
\multicolumn{9}{l}{\textit{Multi-Agent Video Games}} \\
LingBot-World~2.0~\citep{gao2026infinite} & 1256.6 & 0.322 & 13.25 & 0.490 & 0.146 & 0.553 & 0.57 & 5.093 \\
MultiWorld (1-view)$^{a}$~\citep{wu2026multiworld} & 2661.4 & 0.719 & 10.40 & 0.125 & 0.163 & 0.548 & 2.67 & 4.059 \\
Incantation~\citep{zhu2026incantation} & 1989.5 & 0.593 & 10.45 & 0.226 & 0.162 & 0.534 & 8.77 & 4.539 \\
ActionParty~\citep{pondaven2026actionparty} & 1672.5 & 0.272 & 14.56 & 0.467 & 0.086 & 0.547 & 0.47 & 4.974 \\
WanToFight-style$^b$~\citep{hu2026wantofight} & 563.6 & 0.136 & 17.67 & 0.671 & 0.241 & 0.552 & 0.23 & 5.192 \\
\textbf{\ours} & \textbf{338.5} & \textbf{0.086} & \textbf{20.17} & \textbf{0.743} & \textbf{0.594} & \textbf{0.898} & \textbf{0.15} & \textbf{5.207} \\
\midrule
\multicolumn{9}{l}{\textit{Multi-Robot Manipulation}} \\
LingBot-World~2.0~\citep{gao2026infinite} & 2732.5 & 0.603 & 7.13 & 0.464 & 1.106 & 0.665 & 4.20 & 3.570 \\
MultiWorld (1-view)$^{a}$~\citep{wu2026multiworld} & 1748.9 & 0.276 & 15.36 & 0.797 & 1.099 & 0.357 & 1.90 & 4.054 \\
Incantation~\citep{zhu2026incantation} & 3053.3 & 0.820 & 3.93 & 0.185 & 1.119 & 0.459 & 17.37 & 3.710 \\
ActionParty~\citep{pondaven2026actionparty} & 1661.7 & 0.267 & 15.42 & 0.816 & 1.142 & 0.504 & 1.90 & 4.035 \\
WanToFight-style$^{b}$~\citep{hu2026wantofight} & 402.9 & 0.188 & 17.32 & 0.831 & 0.758 & 0.886 & 0.42 & 4.037 \\
\textbf{\ours} & \textbf{270.2} & \textbf{0.106} & \textbf{20.51} & \textbf{0.871} & \textbf{0.558} & \textbf{0.942} & \textbf{0.42} & \textbf{4.085} \\
\bottomrule
\end{tabular}}
\vspace{-1.5em}
\end{table*}

%% file: sec/5_conclusion.tex
\section{Conclusion}
\label{sec:conclusion}

We presented \ours{}, a multi-agent video world model for heterogeneous control across games and robot manipulation. Its Heterogeneous Action Interface combines Agent-Aware Spatial Modulation, which binds each indexed action stream to its visual agent through language-conditioned queries, with Schema-Guided Action Adaptation, which couples each control's language-defined semantics with its synchronized numerical value. Per-agent cross-attention integrates the two within a shared visual representation. Teacher Forcing for AR Tuning and DMD for Streaming Video transfer the interface unchanged to few-step causal rollout. Across six sources, \ours{} achieves best reconstruction and control and reaches 0.898 and 0.942 agent-attribution accuracy without location, mask, or identity input, while its causal generator runs at 17.70 FPS at $512\times448$.

%% file: sec/X_suppl.tex
\clearpage
\setcounter{page}{1}
\maketitlesupplementary

\renewcommand{\thepage}{S\arabic{page}}
\setcounter{section}{0}
\setcounter{figure}{0}
\setcounter{table}{0}
\renewcommand{\thefigure}{S\arabic{figure}}
\renewcommand{\thetable}{S\arabic{table}}
\renewcommand{\theHfigure}{S\arabic{figure}}
\renewcommand{\theHtable}{S\arabic{table}}

\appendix

\startcontents[appendix]
\section*{Appendix}
\setcounter{tocdepth}{2}
\printcontents[appendix]{}{1}{}

\noindent\textcolor{Red}{\textbf{Supplementary video.} Please see the accompanying \texttt{demo.mp4} for additional video results.}

\raggedbottom
\section{Additional Related Work}
\label{sec:additional_related_work}

\noindent\textbf{Structured conditioning in video diffusion.}
Video diffusion models provide temporally coherent generative priors~\citep{ho2022video}, while diffusion transformers and open video foundation models offer scalable token backbones~\citep{peebles2023scalable,ma2025followyourmotion, wang2024cove, wang2026liveedit, ma2026livelight, liu2026opsd, wang2024taming, feng2025dit4edit, qiu2024tfb, qiu2025duet, qiu2025DBLoss, zhang2026zo3t, zhang2026kvpo, xu2026disco, xu2025clgc, song2026vista, song2026streamingeffect, wan2025wan, liu2026omniverifier, liu2026vidforensics, huang2026unityvideo}. Control signals are commonly injected through concatenation, adaptive modulation, or cross-attention. Our distinction is representational rather than merely architectural: a padded action vector is decomposed into per-dimension control tokens, each token carries a task-specific semantic description, and its time-varying scalar modulates rather than replaces that semantic representation. Agent-Aware Spatial Modulation then keeps each stream's provenance intact until cross-attention.

\noindent\textbf{Causal video generation and diffusion distillation.}
Streaming video generation restricts attention to past chunks and caches their keys and values. Self Forcing reduces exposure bias by training on autoregressive model rollouts~\citep{huang2026self, zhuang2026self, su2026sgf+}; Causal Forcing motivates causal teacher initialization before few-step distillation~\citep{zhu2026causal}; and distribution matching distillation expresses the generator update through real- and fake-distribution scores~\citep{yin2024one}. Recent multi-agent systems also combine bidirectional teachers, causal students, and rollout-aware distillation~\citep{savva2026solaris,liu2026gamma,hu2026wantofight}. We adopt this training paradigm to carry the complete HAI into efficient rollout while keeping physical action intervals aligned.

\section{Method Details}
\label{sec:method_details}

\subsection{Problem Formulation}
\label{sec:formulation}

We follow the notation of Sec.~\mainref{sec:method}: $N_{\mathrm{ag}}$ agents with ordered action streams $\mathcal{A}=\{A^{(i)}\}_{i=1}^{N_{\mathrm{ag}}}$, $A^{(i)}=[a_{\tau,j}^{(i)}]\in\mathbb{R}^{T\times C}$, a control schema $S=\{s_j\}_{j=1}^{C}$, and agent descriptions $G=\{g_i\}_{i=1}^{N_{\mathrm{ag}}}$. The interface enforces
\begin{equation}
\begin{gathered}
a_{\tau,j}^{(i)}\longleftrightarrow (g_i,s_j),\\
i\in\{1,\ldots,N_{\mathrm{ag}}\},\quad j\in\{1,\ldots,C\},
\end{gathered}
\label{eq:interface_binding}
\end{equation}
so each numerical control dimension is paired with an agent identity and a task-specific semantic description; the controls of every source are padded to a common width of $C=12$, while $S$ preserves source-specific control semantics.

Let $z_0$ be the VAE encoding of $I_0$ and $z^{1:F-1}$ the future video latents over $F$ latent frames in total. The DiT backbone patchifies these latents into visual tokens $X\in\mathbb{R}^{FN\times D}$, where each latent frame contributes $N$ spatial tokens of hidden dimension $D$. The desired conditional distribution is
\begin{equation}
p_\theta(z^{1:F-1}\mid z_0,c,\mathcal{A},S,G).
\label{eq:objective}
\end{equation}
The bidirectional teacher models all future tokens jointly. The causal generator instead partitions future latents into chunks $\mathcal{B}_1,\ldots,\mathcal{B}_K$ and factorizes
\begin{equation}
p_\theta(z^{1:F-1}\mid z_0,\mathcal{C})
=\prod_{b=1}^{K}p_\theta(z^{\mathcal{B}_b}\mid z^{<\mathcal{B}_b},z_0,\mathcal{C}),
\label{eq:causal_factorization}
\end{equation}
where $\mathcal{C}=(c,\mathcal{A},S,G)$.
Eq.~\ref{eq:interface_binding} makes both agent identity and control semantics explicit; the main-text Method realizes this interface and keeps it under causal rollout.

\subsection{Frame--Latent--Action Alignment}
\label{sec:alignment_details}

The Wan VAE uses temporal compression ratio $r=4$ with a separately encoded first frame. Action sequences are truncated or padded to $T_{\mathrm{need}}=(F-1)r+1$. For a window of $L$ latent frames, latent frame zero repeats the first action $rL$ times, while future latent $k\geq1$ receives
\begin{equation}
\{a_{\tau,j}^{(i)}:\ \tau_k\le\tau\le\tau_k+rL-1\},
\qquad \tau_k=1+r(k-1),
\label{eq:action_start}
\end{equation}
with final-value padding at the sequence boundary; for the primary setting $L=1$ this reduces to the interval $\mathcal{I}_k$ of Eq.~\mainref{eq:causal_action_memory}. The time and control-dimension axes are retained until each local window is formed and then flattened for action attention.

For an 81-frame clip, raw indices $0{:}80$ map to 21 latent frames. Future latent $k\in\{1,\ldots,20\}$ uses actions beginning at $1+4(k-1)$, and chunk $b\in\{1,\ldots,5\}$ contains future latents $4(b-1)+1{:}4b$. During causal teacher forcing, the clean-context/noisy-target shift advances the action interval by one latent. During DMD training, the global latent index is applied not only to the generator's rollout but also to both score models, so all three read the same physical interval. Self-forced score evaluation restores the clean initial latent and excludes it from both DMD and fake-score losses.

\subsection{Causal Condition Propagation}
\label{sec:causal_condition_details}

In the causal generator, the HAI operates only on the real video tokens of the target chunks and returns zero residuals for sequence-padding tokens. Under the clean-context/noisy-target layout used by teacher forcing, actions are injected only into the noisy target half; the clean history remains an unmodified context. During rollout, the clean initial latent is cached without action injection before subsequent target chunks receive their aligned streams. This separation prevents context tokens from being optimized as future action responses.

Checkpoint transfer is performed only between networks with matching HAI architectures and conditioning dimensions.

\subsection{Causal Training Objectives}
\label{sec:causal_training_details}

\paragraph{Foundational Ability Acquisition.}
For clean future latents $z^{1:F-1}$, Gaussian noise $\epsilon$, and sampled time $t$, let $z_t^{1:F-1}=\alpha_t z^{1:F-1}+\sigma_t\epsilon$. The bidirectional model predicts the velocity target $v^*(z^{1:F-1},\epsilon,t)$ under the initial latent, scene, and action conditions:
\begin{equation}
\mathcal{L}_{\mathrm{bi}}=
\mathbb{E}\left[w(t)\left\|v_{\theta_b}(z_t^{1:F-1},z_0,t,\mathcal{C})-v^*\right\|_2^2\right].
\label{eq:bidirectional_loss}
\end{equation}
It provides both the action-conditioned prior and initialization for the causal generator and score networks.

\paragraph{Teacher Forcing for AR Tuning.}
Given the $F=21$ clean latents $z^{0:F-1}$, the training pair uses $z^{0:F-2}$ as context and $z^{1:F-1}$ as targets. The initial frame is conditioning only and is excluded from the loss. We add noise to the $F-1=20$ targets, with every group of four latents sharing a timestep to match chunks:
\begin{equation}
\mathcal{L}_{\mathrm{TF}}=
\mathbb{E}\left[w(t)\left\|v_{\theta_c}(z_t^{1:F-1},z^{0:F-2},t,\mathcal{C}_{\mathrm{shift}})-v^*\right\|_2^2\right].
\label{eq:teacher_forcing_loss}
\end{equation}
Here $\mathcal{C}_{\mathrm{shift}}$ applies the four-raw-frame offset induced by the context--target shift.

\paragraph{DMD for Streaming Video.}
The trainable causal generator $\mathcal{G}_\theta$, which unrolls the causal network $v_{\theta_c}$ with the four-step schedule, initializes its KV cache from the clean initial latent and generates five future chunks from noise without reading future ground-truth latents. The real score $\mu_{\mathrm{real}}$ and the fake score $\mu_{\mathrm{fake},\phi}$ of the main paper receive the same initial frame, scene text, control schema, agent descriptions, and cropped action window as the generator. For $\hat z=\mathcal{G}_\theta(z_0,\mathcal{C})$ and its noised version $\hat z_t$, we normalize the unnormalized distribution-matching gradient $\delta_{\mathrm{DMD}}$ of the main paper as
\begin{equation}
\bar\delta_{\mathrm{DMD}}=
\frac{\delta_{\mathrm{DMD}}}
{\operatorname{mean}\!\left(\left|\hat z-\mu_{\mathrm{real}}(\hat z_t,t,\mathcal{C})\right|\right)+\varepsilon},
\end{equation}
with a small constant $\varepsilon$.
We then optimize the generator through stop-gradient regression,
\begin{equation}
\mathcal{L}_{\mathrm{DMD}}=
\frac{1}{2}\left\|\hat z-
\operatorname{sg}(\hat z-\bar\delta_{\mathrm{DMD}})\right\|_2^2.
\label{eq:dmd_loss}
\end{equation}
The fake score is alternately trained by flow denoising regression on detached samples from the current generator. The default schedule updates it every iteration and updates the generator once every five iterations. We use conditional-only score passes to avoid an additional unconditional forward.

\section{Dataset Details}
\label{sec:dataset_statistics}

\begin{figure*}[t]
    \centering
    \includegraphics[width=\textwidth]{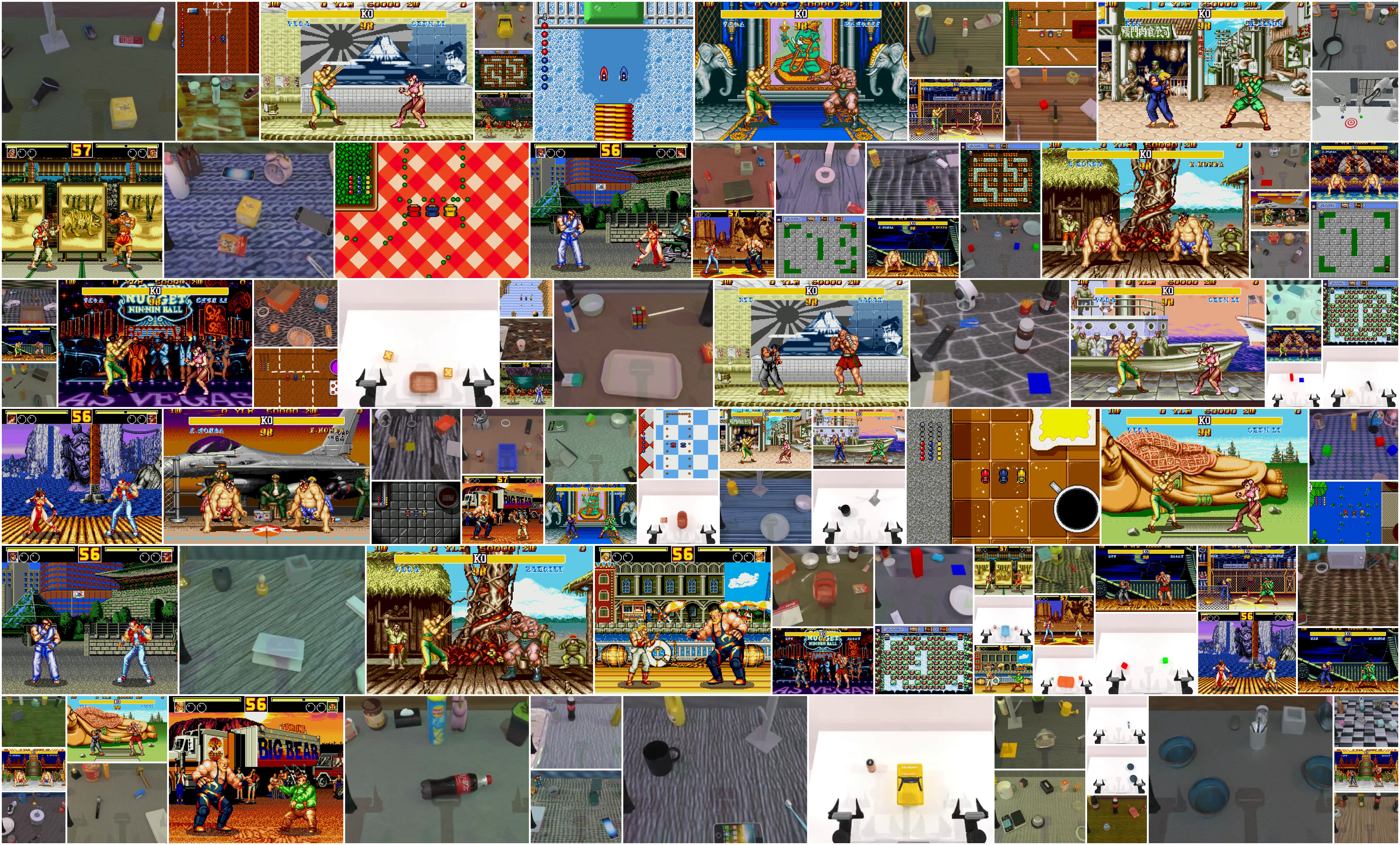}
    \caption{\textbf{Showcase of our dataset.} Frames from the six sources, covering versus fighting, shared-map competition, shared-track racing, and multi-arm manipulation.}
    \label{fig:dataset_showcase}
\end{figure*}

\paragraph{Composition.}
Tab.~\ref{tab:data} summarizes the six sources and Fig.~\ref{fig:dataset_showcase} shows frames from each of them. The dataset contains 17,951 independent multi-agent trajectories, 119.54 hours of raw video, with two to four concurrently controlled agents: the two fighting games and RoboTwin are two-agent, Mega Bomberman and Micro Machines include scenes with two to four agents, and RoboFactory includes two- and three-arm scenes. Every trajectory pairs a shared visual observation with one synchronized action stream per agent, so the number of streams equals the number of agents in the scene.

\paragraph{Sources.}
The game portion is collected through the retro game platform~\citep{stable-retro,nichol2018retro}. These games cover versus fighting, shared-map competition, and shared-track racing. Each trajectory begins from a curated emulator state and records synchronized visual observations and per-agent controller traces under scripted policies at the native 60\,fps. The policy drives each agent's controller independently: it samples one allowed button or a no-op from a game-specific distribution, holds it for a fixed number of frames, and then releases it. We exclude menus, transitions, and incomplete rollouts, yielding 12,860 trajectories of 30\,s each.

The robot portion combines RoboTwin~\citep{mu2024robotwin} and RoboFactory~\citep{qin2025robofactory}. RoboTwin contributes 2,532 episodes across 50 multi-arm tasks: 280 self-collected simulations of 14 tasks at 30\,fps and 2,252 augmented episodes covering all 50 tasks from the randomized LeRobot~\citep{cadene2026lerobot} collection at 50\,fps. RoboFactory contributes 2,559 episodes of cooperative cube stacking at 30\,fps, retaining both successful and unsuccessful trajectories and including two- and three-arm scenes. In both sources, concurrent continuous action streams are aligned with a shared visual observation over the full episode, yielding 5,091 robot trajectories.

\paragraph{Scale and training windows.}
The raw recordings total 24.71\,M frames (games 23.15\,M, 107.17\,h; robots 1.56\,M, 12.37\,h), or 119.54 hours at the native frame rates (Tab.~\ref{tab:data}). Individual trajectories are much longer than a training sample, whereas the model consumes 81-frame windows (Tab.~\ref{tab:training_configuration}). Because generation is conditioned on an initial frame, a window may start at any frame of a trajectory, with its first frame serving as the conditioning image.

\paragraph{Benchmark.}
The 108-clip benchmark of the main paper is kept separate from the training data: it takes one 81-frame clip from each of 108 trajectories that are never used for training, and contains 78 game and 30 robot clips.

\input{tab/dataset_statistics}

\section{Implementation Details}
\label{sec:implementation}

\subsection{Backbone and Conditioner Placement}
\label{sec:backbone_details}

Beyond the placement described in the main paper, the HAI leaves the pretrained conditioning pathways of Wan2.2-TI2V-5B~\citep{wan2025wan} unchanged. The initial image latent is fused into the 48-channel video representation, and a separated timestep assigns zero to this clean latent and the sampled diffusion timestep to future latents; scene text retains the pretrained text cross-attention, diffusion time continues to modulate AdaLN, and video positions retain 3D RoPE. The bidirectional teacher, the causal generator, and both score networks share the same HAI placement and structure.

Control-schema and agent-description embeddings are precomputed with the frozen text encoder and reused throughout training; the pooling operator $\operatorname{Pool}$ of the main paper averages the encoder outputs over the non-padding tokens of each description.

\subsection{Training and Inference Configuration}
\label{sec:training_configuration}

Tab.~\ref{tab:training_configuration} collects the backbone, data, and optimization settings of the three stages in Fig.~\mainref{fig:overview}, including those stated in the main paper. Beyond them, the primary configuration uses description-based, value-modulated control tokens, and all stages share a flow-matching parameterization with 1000 training timesteps and a timestep shift of 5.0. Foundational Ability Acquisition uses AdamW with $\beta=(0.9,0.999)$ and a constant learning rate, whereas the two causal stages use $\beta=(0,0.999)$ and clip gradient norms at 10. DMD for Streaming Video maintains an EMA of the generator with decay 0.99 from iteration 200 onward, and the EMA weights are used at inference.

\begin{table*}[t]
\centering
\caption{\textbf{Training and inference configuration of \ours.} Stages 2 and 3 are trained with FSDP, and Stage 3 samples at timesteps $[1000,750,500,250]$ with the EMA weights; ``--'' marks settings that do not apply.}
\label{tab:training_configuration}
\small
\renewcommand{\arraystretch}{1.12}
\begin{tabularx}{\textwidth}{@{}l*{3}{>{\centering\arraybackslash}X}@{}}
\toprule
\multicolumn{4}{@{}l}{\textit{Model and data}} \\
\midrule
Backbone & \multicolumn{3}{l}{Wan2.2-TI2V-5B; 30 blocks, width 3072, 24 heads; VAE $4\times16\times16$} \\
Trainable & \multicolumn{3}{l}{Full DiT and HAI; umT5 encoder and VAE frozen} \\
Clip & \multicolumn{3}{l}{81 frames at $512\times448$, 16\,fps; 1 conditioning and 20 future latents} \\
Frame sampling & \multicolumn{3}{l}{Games: consecutive native frames; robots: uniform over the episode} \\
Actions & \multicolumn{3}{l}{12 description-modulated controls per agent; one-latent window} \\
Conditioner & \multicolumn{3}{l}{Text projections $4096\rightarrow1024\rightarrow3072$; value/gate/delta bottleneck 768} \\
\midrule
\textit{Training and inference} & \textbf{Stage 1} & \textbf{Stage 2} & \textbf{Stage 3} \\
 & Foundation & Teacher Forcing & DMD \\
\cmidrule(l){2-4}
Optimizer & \multicolumn{3}{c}{AdamW, weight decay 0.01, BF16} \\
Training steps & 10k & 6k & 700 \\
Global batch size & 16 & 24 & 24 \\
Learning rate & $10^{-4}$ & $2\times10^{-6}$ & $2\times10^{-6}$ \\
Fake-score LR & -- & -- & $4\times10^{-7}$ \\
AdamW $\beta$ & $(0.9, 0.999)$ & $(0, 0.999)$ & $(0, 0.999)$ \\
Gradient clipping & -- & 10 & 10 \\
Generator update & every step & every step & every 5 steps \\
Generator EMA & -- & -- & 0.99 from step 200 \\
Denoising steps & 50 & 50 & 4 \\
Classifier-free guidance & 5.0 & 5.0 & -- \\
\bottomrule
\end{tabularx}
\end{table*}

\section{Evaluation Details}
\label{sec:evaluation_details}

\subsection{Evaluation Protocol}
\label{sec:eval_protocol}

Both evaluation tracks of Sec.~\mainref{sec:main_results} ask the same three questions. \textbf{First}, does Agent-Aware Spatial Modulation assign each indexed action stream to the correct visual agent in the jointly generated scene? \textbf{Second}, does Schema-Guided Action Adaptation interpret heterogeneous controls consistently across games and embodied tasks? \textbf{Third}, how much control and visual quality survive causal transfer, and what latency is gained? The first two questions evaluate the complementary components of the HAI.

Every reported method is evaluated on the benchmark of App.~\ref{sec:dataset_statistics} with one 81-frame prediction at seed 0 per condition. Frame-paired metrics exclude the conditioning frame and average 16 uniformly sampled frames from frames 1--80. Pixel metrics use $256^2$ inputs, while FVD uses canonical Kinetics-400 I3D pre-softmax features at $224^2$. Since every clip starts from a fixed initial frame and a fixed set of action streams in a deterministic environment, frame-paired errors measure deviation from the commanded reference future; the control metrics separately assess execution and attribution. Causal efficiency is evaluated separately because published throughput values use different backbones, resolutions, hardware, and decoding pipelines. Architecture ablations use the same data, optimization budget, and evaluation seed.

\subsection{Baseline Configurations}
\label{sec:baselines}

All five baselines are evaluated on the same clips and protocol. LingBot-World~2.0, MultiWorld, and Incantation do not release training code, so we run their released checkpoints with their original model configurations and conditioning interfaces, without further training. ActionParty does not release trained checkpoints, so we train it with the same budget on our dataset. The WanToFight code is not publicly available, so we train our own reimplementation on the same mixture and denote it WanToFight-style.
\begin{itemize}[leftmargin=*,nosep]
    \item \textbf{LingBot-World~2.0}~\citep{gao2026infinite} serializes all agents into one time-varying prompt stream, including explicit continuous robot values.
    \item \textbf{MultiWorld}~\citep{wu2026multiworld} retains its Multi-Agent Condition Module, while the Global State Encoder receives only the same single initial view available to other methods.
    \item \textbf{Incantation}~\citep{zhu2026incantation} uses per-frame, per-entity language actions, with robot values represented explicitly in text.
    \item \textbf{ActionParty}~\citep{pondaven2026actionparty} jointly predicts persistent subject-state tokens and video tokens. It receives initial agent locations as required by its binding mechanism; we extend its discrete action projection to continuous robot values and train the resulting model.
    \item \textbf{WanToFight-style}~\citep{hu2026wantofight} combines per-agent locally causal action injection with first-frame identity crops and reference-conditioned Player Association. The original code is not publicly available; our reported results are obtained by training this reproduction.
    \item \textbf{\ours} uses per-agent schema-numerical controls without locations, masks, boxes, or identity crops.
\end{itemize}

\subsection{Metrics}
\label{sec:control_metrics}

\noindent\textbf{Reconstruction quality.}
FVD~\citep{unterthiner2018towards}, LPIPS~\citep{zhang2018unreasonable}, PSNR, and SSIM follow the sampling protocol of App.~\ref{sec:eval_protocol}: FVD measures distributional similarity using I3D features, whereas the three frame-paired metrics measure perceptual, pixel, and structural agreement with the ground-truth future.

\noindent\textbf{Motion and appearance diagnostics.}
Motion Magnitude Error (MME) is $\lvert \rho(\mathrm{gen})-\rho(\mathrm{GT})\rvert$, where $\rho(\cdot)$ is the mean RAFT~\citep{teed2020raft} flow magnitude over the same 16-frame $256^2$ grid; lower values indicate closer aggregate motion energy. Aesthetic uses the LAION improved predictor with CLIP ViT-L/14.

\noindent\textbf{Control and attribution.}
Following MultiWorld~\citep{wu2026multiworld}, the inverse dynamics evaluator of the main paper is trained only on real training clips and never on generated videos; it consists of one frozen convolutional network with about 2.9M parameters per data source. For each action frame, the network reads the surrounding RGB frames resized to $128^2$, using a five-frame window with offsets $\{-6,-3,0,3,6\}$ for games and a three-frame window $\{-1,0,1\}$ for robots, and jointly predicts the controls of both agents. It receives no agent identity or location, so it can tell the agents apart only from the pixels. Predictions cover frames 1--80 of each clip, excluding the conditioning frame. For games, it predicts movement and interaction events for each agent, and Game Act-F1 is the macro F1 across valid events and agents. For robots, it predicts the continuous control trajectory. Let $\mathcal{V}$ denote the valid, non-padding control dimensions, which are the only ones scored, and let $\sigma_j$ be the standard deviation of dimension $j$ on the real training set. We compute
\begin{equation}
\operatorname{NRMSE}
=
\sqrt{
\frac{1}{T|\mathcal{V}|}
\sum_{\tau=1}^{T}\sum_{j\in\mathcal{V}}
\left(
\frac{\widehat{a}_{\tau,j}-a_{\tau,j}}{\sigma_j+\varepsilon}
\right)^2
},
\label{eq:robot_nrmse}
\end{equation}
where $\widehat{a}_{\tau,j}$ is the recovered control, $\varepsilon$ is a small constant, and lower values indicate more accurate control recovery.

Attr-Acc is a pairwise swap test on the same generated videos and evaluator predictions, so it requires no additional renders and is computed identically for every method in Tab.~\mainref{tab:main}. For a clip with $N_{\mathrm{ag}}$ agents, let $\mathbf{u}^{(i)}_{\tau}$ and $\widehat{\mathbf{u}}^{(i)}_{\tau}$ be the commanded and recovered controls of agent $i$ on $\mathcal{V}$, centered by subtracting the mean of the same source, agent, and dimension on the evaluator's training set. Robot dimensions are further divided by the corresponding training-set standard deviation, whereas game dimensions already lie in $[0,1]$ and are not rescaled. At frame $\tau$, an unordered agent pair $(i,i')$ is scored only if their commanded streams differ on at least one dimension in $\mathcal{V}$, since exchanging identical commands is uninformative. We compare the commanded assignment with the one in which only agents $i$ and $i'$ are exchanged,
\begin{equation}
\begin{aligned}
E^{\mathrm{c}}_{\tau,ii'}&=\sum_{k\in\{i,i'\}}\bigl\lVert\widehat{\mathbf{u}}^{(k)}_{\tau}-\mathbf{u}^{(k)}_{\tau}\bigr\rVert_2^2,\\
E^{\mathrm{s}}_{\tau,ii'}&=\bigl\lVert\widehat{\mathbf{u}}^{(i')}_{\tau}-\mathbf{u}^{(i)}_{\tau}\bigr\rVert_2^2+\bigl\lVert\widehat{\mathbf{u}}^{(i)}_{\tau}-\mathbf{u}^{(i')}_{\tau}\bigr\rVert_2^2,
\end{aligned}
\end{equation}
where the remaining agents contribute equally to both assignments and are omitted. The pair scores 1 if $E^{\mathrm{c}}_{\tau,ii'}<E^{\mathrm{s}}_{\tau,ii'}$, 0.5 on a tie, and 0 otherwise. It is therefore correct when the controls recovered from the pixels match the commands better in the commanded order than with the two agents exchanged, that is, when each of the two streams drives its intended agent rather than the other one. A frame is scored if it has at least one scored pair, and its score is the mean over these pairs; with $N_{\mathrm{ag}}{=}2$ there is a single pair and the test reduces to exchanging the two agents. The centering fixes the chance level at 0.5 for any $N_{\mathrm{ag}}$: a predictor that ignores the video and outputs each agent's training mean is zero for every agent after centering, so $E^{\mathrm{c}}_{\tau,ii'}=E^{\mathrm{s}}_{\tau,ii'}$ for every pair. Frame scores are averaged within each clip, then over the clips of each source, and finally over sources without weighting, four for games and two for robots. Averaging within clips first prevents clips with many scored frames from dominating, and the source-level mean gives every source equal weight.

\section{Additional Results}
\label{sec:additional_results}

\subsection{Three-Stage Causal Transfer}
\label{sec:causal_stage_results}

Tab.~\ref{tab:causal_stages} summarizes the generation settings of the three stages together with their control, attribution, and throughput results. Stage 2 runs slower than Stage 1 despite its causal attention: it repeats 50 denoising steps with classifier-free guidance (CFG) sequentially for each of the five four-latent chunks and refreshes the KV cache after every chunk, whereas the bidirectional teacher denoises all 21 latents jointly at each step. These many small, sequential forward passes use the GPU less efficiently than the teacher's large joint passes. Stage 3 reduces each chunk to four denoising steps without CFG and reaches 17.70 FPS, $10.7\times$ faster than Stage 2.
\input{tab/causal_stages}

\subsection{Detailed Ablations}
\label{sec:detailed_ablations}

Tab.~\ref{tab:detailed_ablation} extends the main-text ablation along two axes: the construction of the schema-guided control tokens and the temporal alignment of App.~\ref{sec:alignment_details}.

\input{tab/sgaa_variants}

\noindent\textbf{Schema-Guided Action Adaptation.}
SGAA combines three factors: a schema description that identifies each control, a separate token for each control dimension, and affine modulation of that token by the control value. Tab.~\ref{tab:sgaa_variants} summarizes the variants that separate these factors. Every variant changes only how the control tokens are constructed and keeps Agent-Aware Spatial Modulation, per-agent cross-attention, the one-latent action window, and the time and agent embeddings of Eq.~\mainref{eq:action_token}.

The main-text w/o SGAA variant replaces the module with a conventional flat action encoder that maps the full action vector of each frame to a single token. It removes all three factors at once and reduces the action memory of each latent from $rC$ to $r$ tokens. The remaining variants keep the per-dimension tokens and remove one factor at a time. The w/o value modulation variant keeps the schema token but adds a value embedding to it instead of modulating it. The w/o schema semantics variant keeps the modulation but replaces the schema embedding with a learned per-dimension embedding, so that each dimension remains distinguishable but carries no language description; comparing it with the full model tests whether the schema description contributes beyond identifying the dimension.

Replacing modulation with addition alone collapses control to the level of removing SGAA in both domains (game Act-F1 0.209 vs.\ 0.225, robot NRMSE 1.071 vs.\ 1.120), and attribution is no better (game Attr-Acc 0.513 vs.\ 0.603, robot Attr-Acc 0.480 vs.\ 0.484), with robot Attr-Acc near the chance level of 0.5. Schema descriptions and per-dimension tokens are therefore insufficient without modulation. This is consistent with the design of Sec.~\mainref{sec:action_conditioner}: addition mixes the value embedding into the schema token, whereas modulation keeps the schema token as the carrier of the control identity and lets the value only rescale and shift it.

Replacing the schema embedding with learned per-dimension IDs isolates the contribution of the language description. On games, this variant falls to the level of removing SGAA (Act-F1 0.230 vs.\ 0.225, Attr-Acc 0.588 vs.\ 0.603), even though it keeps the per-dimension tokens and value modulation of the full model. The four game sources assign different controls to the same index, so an index-level ID cannot distinguish these controls, whereas the schema description can. On robots, learned IDs retain part of the gain over removing SGAA (NRMSE 0.799 vs.\ 1.120, Attr-Acc 0.778 vs.\ 0.484) but remain clearly below the full model (0.558 and 0.942). The schema description therefore contributes beyond identifying each dimension, most strongly on the games, where the same index carries different meanings across sources.

We additionally test whether the trained full model actually reads the schema descriptions. Without retraining, we pair each control dimension with the description of another dimension at inference, keeping the action values, per-dimension tokens, and value modulation unchanged. Control collapses to the level of the reduced variants: game Act-F1 drops to 0.207, and Attr-Acc falls to 0.504 on games and 0.496 on robots, both near chance. The full model therefore relies on the schema embedding to identify each control, and the per-dimension tokens and value modulation cannot produce the commanded behavior once the descriptions are mismatched.

\noindent\textbf{Temporal alignment.}
The action window length $L$ sets how many latent frames of controls each future latent attends to: the main model uses $L{=}1$, so every latent reads exactly the $r{=}4$ raw-time steps compressed into it, while $L{=}2$ and $L{=}4$ extend the window over the following latents. Longer windows dilute the per-latent control signal and degrade all four metrics monotonically, on games (Act-F1 $0.594\rightarrow0.449\rightarrow0.325$, Attr-Acc $0.898\rightarrow0.846\rightarrow0.721$) as well as on robots (NRMSE $0.558\rightarrow0.591\rightarrow0.700$, Attr-Acc $0.942\rightarrow0.937\rightarrow0.926$).

\input{tab/detailed_ablation}

\subsection{Spatial Localization of Agent-Specific Residuals}
\label{sec:residual_field}

\begin{figure*}[t]
    \centering
    \begin{tikzpicture}
        \node[anchor=south west,inner sep=0] (residualmaps) at (0,0) {\includegraphics[width=0.9\textwidth]{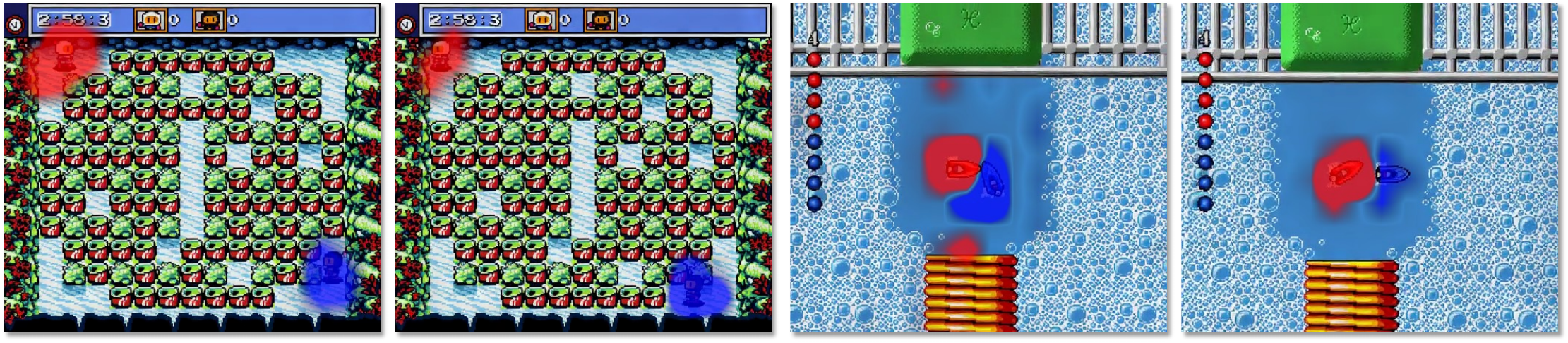}};
        \begin{scope}[x={(residualmaps.south east)},y={(residualmaps.north west)}]
            \node[anchor=north west,fill=white,fill opacity=0.8,text opacity=1,inner sep=1pt,font=\scriptsize\bfseries] at (0.01,0.98) {(a)};
            \node[anchor=north west,fill=white,fill opacity=0.8,text opacity=1,inner sep=1pt,font=\scriptsize\bfseries] at (0.26,0.98) {(b)};
            \node[anchor=north west,fill=white,fill opacity=0.8,text opacity=1,inner sep=1pt,font=\scriptsize\bfseries] at (0.51,0.98) {(c)};
            \node[anchor=north west,fill=white,fill opacity=0.8,text opacity=1,inner sep=1pt,font=\scriptsize\bfseries] at (0.76,0.98) {(d)};
        \end{scope}
    \end{tikzpicture}
    \caption{\textbf{Spatial localization of agent-specific control residuals.} Panels (a)--(d) show representative spatial residual-magnitude heatmaps from the agent-specific branches. Red and blue denote agents 1 and 2, respectively, and higher intensity indicates a larger response. Across the examples, the high-response regions concentrate around the corresponding visual agents.}
    \label{fig:residual_field}
\end{figure*}

Fig.~\ref{fig:residual_field} provides a complementary spatial diagnostic of Agent-Aware Spatial Modulation: each agent-specific residual-magnitude heatmap concentrates around the corresponding visual agent, consistent with the intended stream-to-agent binding.

\subsection{Additional Qualitative Comparisons}
\label{sec:more_qualitative}

\begin{figure*}[t]
    \centering
    \includegraphics[width=\textwidth]{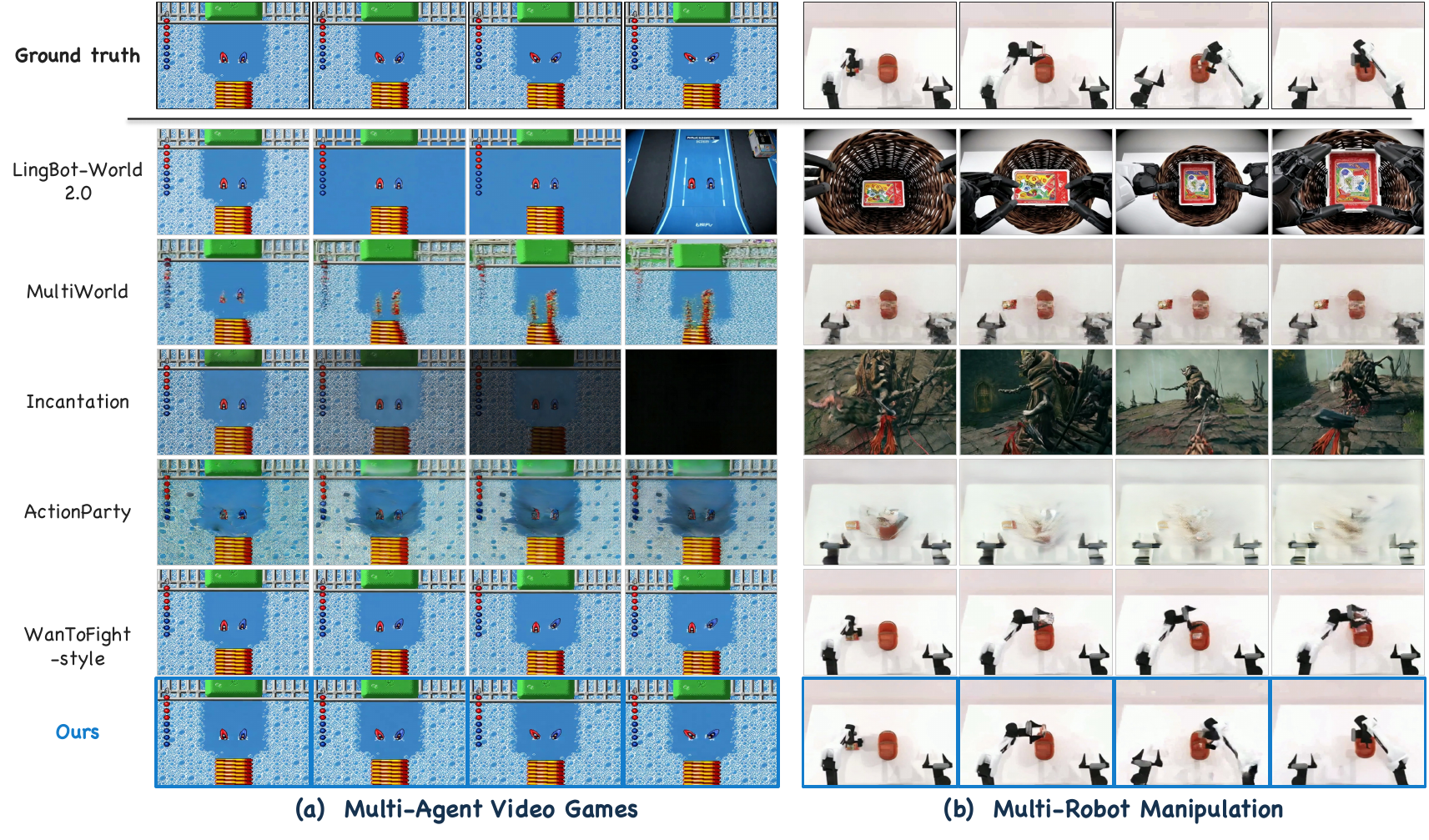}
    \caption{\textbf{Additional qualitative comparison (set 1).} Same layout and methods as Fig.~\mainref{fig:qualitative}.}
    \label{fig:qual_supp_1}
\end{figure*}

\begin{figure*}[t]
    \centering
    \includegraphics[width=\textwidth]{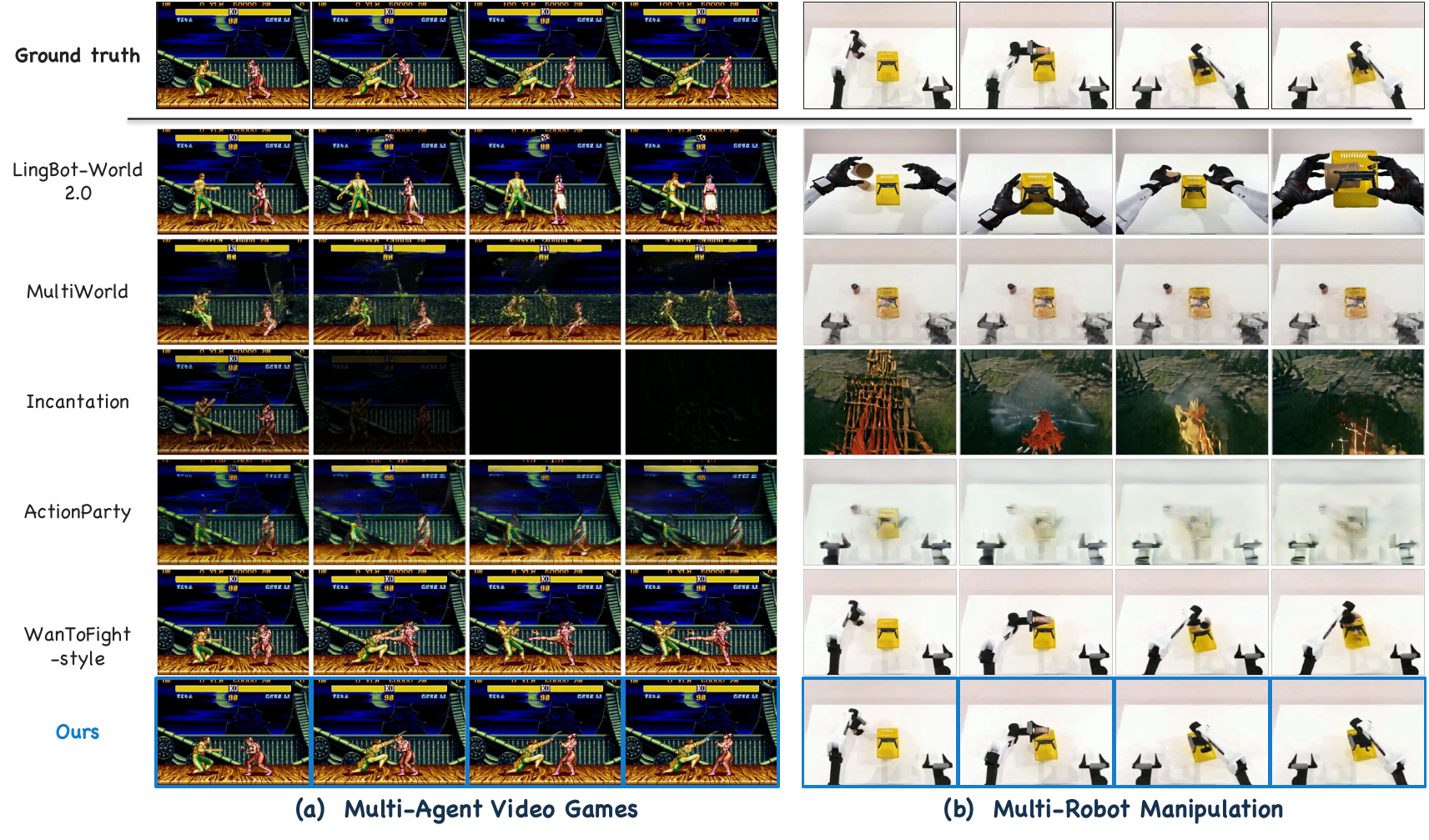}
    \caption{\textbf{Additional qualitative comparison (set 2).} Same layout and methods as Fig.~\mainref{fig:qualitative}.}
    \label{fig:qual_supp_2}
\end{figure*}

Fig.~\ref{fig:qual_supp_1} and Fig.~\ref{fig:qual_supp_2} add two pairs of cases with the same methods and sampling protocol as Fig.~\mainref{fig:qualitative}: a Mega Bomberman round and a Street Fighter~II round, each paired with a RoboTwin scene in which one arm places an object into a basket. They confirm that the failure modes of the main paper are not specific to the two scenes shown there. 
First, the scene drift of LingBot-World~2.0 can be abrupt: In Fig.~\ref{fig:qual_supp_1}, the generated frames gradually turn black and then transitions to an unrelated scene. However, in both robot cases, it started to break down from the first sampling frame.
Second, the game-track failures of Incantation grow with time, from a darkened board to black frames, while MultiWorld and ActionParty keep the board and the table but smear the players and arms. 
Although WanToFight maintains the scene and completes the action, there is a mismatch between the arm and the action. \ours{} matches the target arm pose more closely at each sampled frame.

\subsection{Additional \ours{} Showcases}
\label{sec:ours_showcases}

\begin{figure*}[t]
    \centering
    \includegraphics[width=0.98\textwidth,height=0.92\textheight,keepaspectratio]{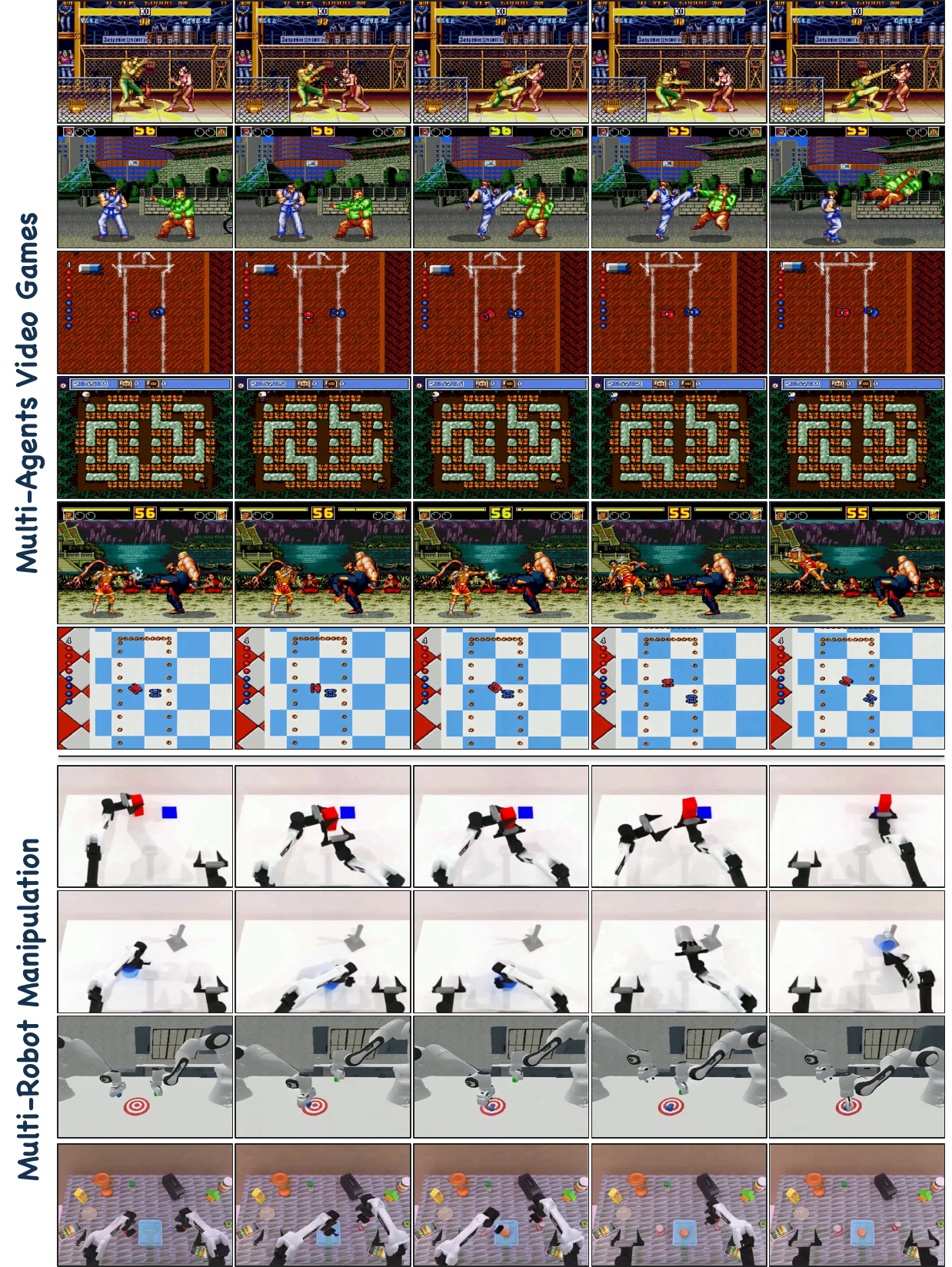}
    \caption{\textbf{Additional \ours{} showcase.}}
    \label{fig:ours_showcase}
\end{figure*}

Fig.~\ref{fig:ours_showcase} presents additional generations from \ours{} across the game and robot domains. Each row shows one evaluation condition, and columns show the generated rollout at successive times.

\label{sec:multi_agent_showcase}

\begin{figure*}[t]
    \centering
    \includegraphics[width=0.82\textwidth]{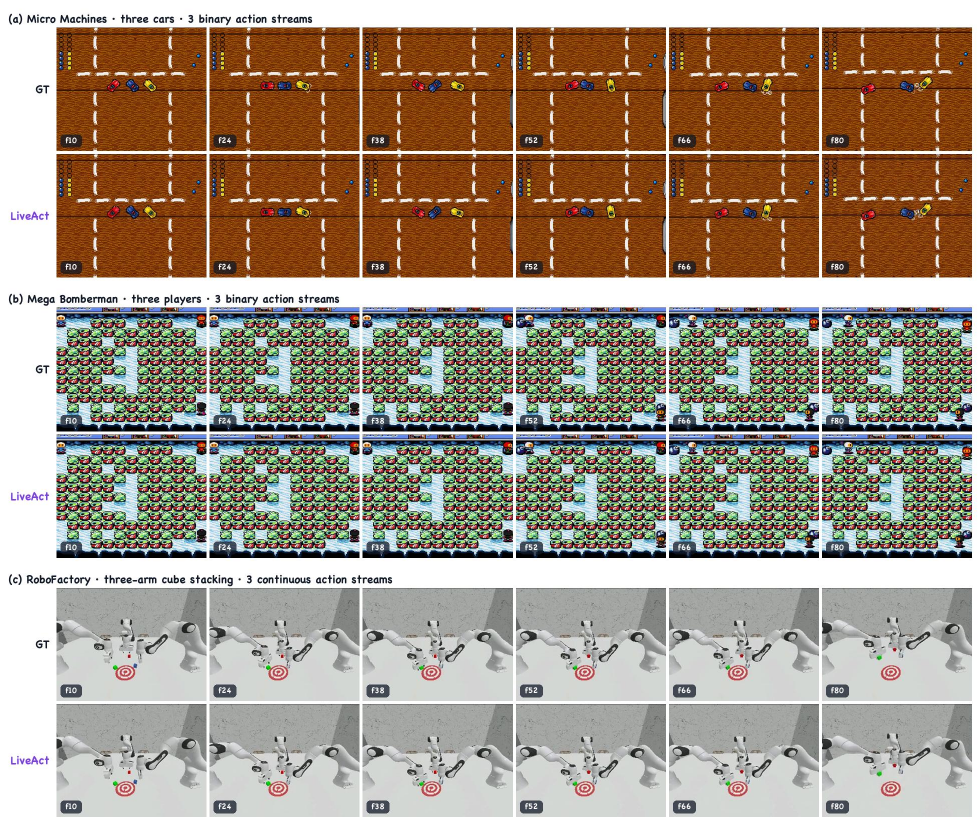}
    \caption{\textbf{Showcase beyond two agents.} Scenes with more than two concurrently controlled agents. For each scene, the top row shows the ground truth and the bottom row the \ours{} generation from the same initial frame and per-agent action streams.}
    \label{fig:multi_agent_showcase}
\end{figure*}

The interface is not tied to a fixed number of agents: the dataset contains scenes with two to four concurrently controlled agents, and each additional agent only adds one agent description, one action stream, and one per-agent cross-attention residual, without any architectural change. Fig.~\ref{fig:multi_agent_showcase} shows scenes with more than two agents: a Micro Machines race and a Mega Bomberman round with binary controls, and a RoboFactory cube-stacking episode with continuous controls.

\subsection{Long-Horizon Rollout Showcase}
\label{sec:long_horizon_showcases}

\begin{figure*}[t]
    \centering
    \includegraphics[width=0.82\textwidth]{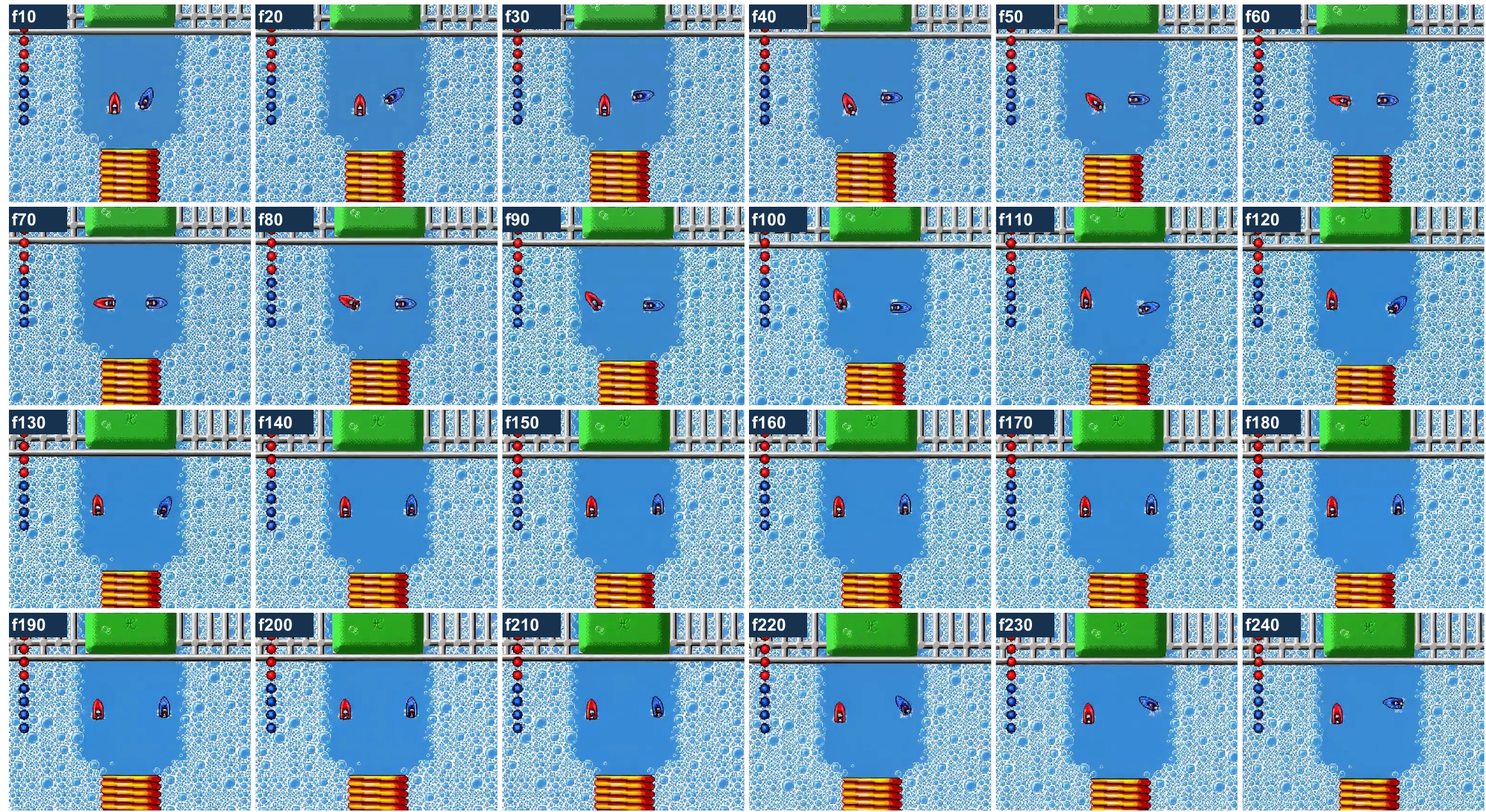}
    \caption{\textbf{Single-case long-horizon rollout as a dense frame matrix.} We show 24 generated frames from the same 241-frame rollout, sampled every ten frames and ordered from left to right and top to bottom.}
    \label{fig:long_video_cases}
\end{figure*}

Fig.~\ref{fig:long_video_cases} shows an additional dynamic case that exceeds the 81-frame training and evaluation window used for quantitative assessment.

%% file: tab/dataset_statistics.tex
\begin{table*}[t]
\centering
\caption{\textbf{Dataset statistics for four multi-agent games and two multi-robot manipulation sources.} Frame counts and durations refer to the raw recordings at their native frame rates (RoboTwin: 30\,fps for self-collected and 50\,fps for augmented episodes). Training and evaluation use 81-frame clips drawn from these trajectories.}
\label{tab:data}
\setlength{\tabcolsep}{4pt}
\resizebox{\textwidth}{!}{
\begin{tabular}{lccccccc}
\toprule
\midrule

& Street Fighter~II & Fatal Fury~2 & Mega Bomberman & Micro Machines & RoboTwin & RoboFactory & \textbf{Total} \\
\midrule
\textbf{Trajectories} & 5460 & 3200 & 2050 & 2150 & 2532 & 2559 & \textbf{17951} \\

\textbf{Agents} & 2 & 2 & 2--4 & 2--4 & 2 & 2--3 & \textbf{2--4} \\
\textbf{Action type} & binary & binary & binary & binary & continuous & continuous & \textbf{mixed} \\
\textbf{Frame rate (fps)} & 60 & 60 & 60 & 60 & 30 / 50 & 30 & -- \\
\textbf{Raw frames (M)} & 9.83 & 5.76 & 3.69 & 3.87 & 0.65 & 0.91 & \textbf{24.71} \\
\textbf{Raw duration (h)} & 45.50 & 26.67 & 17.08 & 17.92 & 3.93 & 8.44 & \textbf{119.54} \\
\midrule
\bottomrule
\end{tabular}}
\vspace{-1em}
\end{table*}

%% file: tab/causal_stages.tex
\begin{table*}[t]
\centering
\caption{\textbf{Three-stage causal transfer.} Generation settings and results of each stage; throughput is measured at $512\times448$ on a single GPU, and Stage 3 samples without classifier-free guidance (CFG). Best results are bolded.}
\label{tab:causal_stages}
\setlength{\tabcolsep}{6pt}
\resizebox{\textwidth}{!}{
\begin{tabular}{lccc}
\toprule
 & \shortstack{\textbf{Stage 1}\\\textbf{Foundational Ability Acquisition}} & \shortstack{\textbf{Stage 2}\\\textbf{Teacher Forcing for AR Tuning}} & \shortstack{\textbf{Stage 3}\\\textbf{DMD for Streaming Video}} \\
\midrule
Role & Full-context teacher & Causal initialization & Rollout distillation \\
Attention & Bidirectional & Blockwise causal & Blockwise causal \\
Training history & Full video & Clean chunks & Model rollouts \\
Streaming & $\times$ & $\checkmark$ & $\checkmark$ \\
Chunk size & Full sequence & 4 latent frames & 4 latent frames \\
Denoising steps & 50 & 50 & 4 \\
CFG scale & 5.0 & 5.0 & -- \\
\midrule
Game Act-F1 $\uparrow$ & 0.539 & 0.565 & \textbf{0.594} \\
Game Attr-Acc $\uparrow$ & 0.844 & \textbf{0.898} & \textbf{0.898} \\
Robot NRMSE $\downarrow$ & \textbf{0.430} & 0.473 & 0.558 \\
Robot Attr-Acc $\uparrow$ & 0.969 & \textbf{0.970} & 0.942 \\
FPS $\uparrow$ & 2.77 & 1.66 & \textbf{17.70} \\
\bottomrule
\end{tabular}}
\end{table*}

%% file: tab/sgaa_variants.tex
\begin{table*}[t]
\centering
\caption{\textbf{Design of the Schema-Guided Action Adaptation variants.} Each variant changes only how the control tokens of agent $i$ are constructed; the superscript $(i)$ is omitted. All variants keep Agent-Aware Spatial Modulation, per-agent cross-attention, the one-latent action window, and the time and agent embeddings of Eq.~\mainref{eq:action_token}. $u_j$ is the projected schema embedding of Eq.~\mainref{eq:schema_encoding}, $(\gamma_{\tau,j},\beta_{\tau,j})=\operatorname{MLP}_{\mathrm{mod}}(a_{\tau,j})$ as in Eq.~\mainref{eq:modulated_control}, $e_j^{\mathrm{ID}}$ is a learned per-dimension embedding, and $\pi$ is a fixed permutation of the control dimensions.}
\label{tab:sgaa_variants}
\setlength{\tabcolsep}{4pt}
\resizebox{\textwidth}{!}{
\begin{tabular}{lccccl}
\toprule
Variant & Control identity & Tokens per frame & Value fusion & Training & Control token \\
\midrule
Full \ours{} & schema $u_j$ & $C$ & affine modulation & trained & $h_{\tau,j}=u_j\odot(1+\gamma_{\tau,j})+\beta_{\tau,j}$ \\
\quad w/o SGAA & -- & $1$ & flat MLP & trained & $h_{\tau}=\operatorname{MLP}_{\mathrm{flat}}(A_{\tau})$ \\
\quad w/o value modulation & schema $u_j$ & $C$ & addition & trained & $h_{\tau,j}=u_j+\operatorname{MLP}_{\mathrm{val}}(a_{\tau,j})$ \\
\quad w/o schema semantics & learned ID $e_j^{\mathrm{ID}}$ & $C$ & affine modulation & trained & $h_{\tau,j}=e_j^{\mathrm{ID}}\odot(1+\gamma_{\tau,j})+\beta_{\tau,j}$ \\
\quad shuffled schema & schema $u_{\pi(j)}$ & $C$ & affine modulation & inference only & $h_{\tau,j}=u_{\pi(j)}\odot(1+\gamma_{\tau,j})+\beta_{\tau,j}$ \\
\bottomrule
\end{tabular}}
\end{table*}

%% file: tab/detailed_ablation.tex
\begin{table}[t]
\centering
\caption{\textbf{Detailed ablations of Schema-Guided Action Adaptation and temporal alignment.} The SGAA variants are defined in Tab.~\ref{tab:sgaa_variants}. All trained variants use the same data and optimization budget as the full model. $L$ is the action window length in latent frames (App.~\ref{sec:alignment_details}). Best results are bolded.}
\label{tab:detailed_ablation}
\setlength{\tabcolsep}{4pt}
\resizebox{\linewidth}{!}{
\begin{tabular}{lcccc}
\toprule
Variant & Game Act-F1 $\uparrow$ & Game Attr-Acc $\uparrow$ & Robot NRMSE $\downarrow$ & Robot Attr-Acc $\uparrow$ \\
\midrule
Full \ours{} ($L{=}1$) & \textbf{0.594} & \textbf{0.898} & \textbf{0.558} & \textbf{0.942} \\
\addlinespace[2pt]
\multicolumn{5}{l}{\textbf{Schema-Guided Action Adaptation}} \\
\quad w/o SGAA (flat MLP) & 0.225 & 0.603 & 1.120 & 0.484 \\
\quad w/o value modulation (additive value) & 0.209 & 0.513 & 1.071 & 0.480 \\
\quad w/o schema semantics (ID tokens) & 0.230 & 0.588 & 0.799 & 0.778 \\
\quad shuffled schema (inference only) & 0.207 & 0.504 & 1.063 & 0.496 \\
\addlinespace[2pt]
\multicolumn{5}{l}{\textbf{Temporal alignment}} \\
\quad action window $L{=}2$ & 0.449 & 0.846 & 0.591 & 0.937 \\
\quad action window $L{=}4$ & 0.325 & 0.721 & 0.700 & 0.926 \\
\bottomrule
\end{tabular}}
\end{table}